\pdfoutput=1

\documentclass[letterpaper,twocolumn,10pt]{article}
\usepackage{usenix2019_v3}

\usepackage[utf8]{inputenc} 
\usepackage[T1]{fontenc}    
\usepackage{hyperref}       
\usepackage{url}            
\usepackage{booktabs}       
\usepackage{amsfonts}       
\usepackage{nicefrac}       
\usepackage{microtype}      
\usepackage{xspace}
\usepackage{graphicx}
\usepackage{subcaption}
\usepackage{multirow}

\usepackage{algorithm}
\usepackage{algpseudocode}
\usepackage{xcolor}
\usepackage{kotex}
\usepackage[most]{tcolorbox}
\usepackage[normalem]{ulem}
\usepackage{tikz}
\usepackage{makecell}
\usepackage{float}
\usepackage{textcomp}

\newcommand*\circled[1]{\tikz[baseline=(char.base)]{
            \node[shape=circle,draw,inner sep=0.5pt] (char) {#1};}}

\newcommand{\system}[0]{Hippo\xspace}
\newcommand{\hp}[0]{hyper-parameter\xspace}
\newcommand{\hps}[0]{hyper-parameters\xspace}
\newcommand{\Hp}[0]{Hyper-parameter\xspace}

\newcommand{\hpo}[0]{hyper-parameter optimization\xspace}
\newcommand{\Hpo}[0]{Hyper-parameter optimization\xspace}
\newcommand{\HPO}[0]{Hyper-Parameter Optimization\xspace}
\newcommand{\hpoa}[0]{hyper-parameter optimization algorithm\xspace}

\newcommand{\hpoas}[0]{hyper-parameter optimization algorithms\xspace}
\newcommand{\study}[0]{study\xspace}
\newcommand{\studies}[0]{studies\xspace}
\newcommand{\trial}[0]{trial\xspace}
\newcommand{\trials}[0]{trials\xspace}
\newcommand{\Trial}[0]{Trial\xspace}
\newcommand{\Trials}[0]{Trials\xspace}
\newcommand{\stage}[0]{stage\xspace}
\newcommand{\stages}[0]{stages\xspace}
\newcommand{\Stage}[0]{Stage\xspace}
\newcommand{\Stages}[0]{Stages\xspace}
\newcommand{\plan}[0]{search plan\xspace}

\newcommand{\PLan}[0]{Search Plan\xspace}
\newcommand{\plans}[0]{search plans\xspace}

\newcommand{\tree}[0]{stage tree\xspace}
\newcommand{\trees}[0]{stage trees\xspace}

\newcommand{\Trees}[0]{Stage trees\xspace}

\newcommand{\searchspace}{search space\xspace}
\newcommand{\searchspaces}{search spaces\xspace}

\usepackage{listings}

\DeclareFixedFont{\ttb}{T1}{txtt}{bx}{n}{8} 
\DeclareFixedFont{\ttm}{T1}{txtt}{m}{n}{8}

\definecolor{deepblue}{rgb}{0,0,0.5}
\definecolor{deepred}{rgb}{0.6,0,0}
\definecolor{deepgreen}{rgb}{0,0.5,0}
\definecolor{gray}{rgb}{0.33,0.33,0.33}
\newcommand\pythonstyle{\lstset{
		language=Python,
		basicstyle=\ttm,
		commentstyle=\color{deepgreen},
		keywordstyle=\ttb\color{deepblue},
		emphstyle=\ttb\color{deepred},    
		frame=tb,                         
		showstringspaces=false,            %
		numbers=left,
}}

\lstnewenvironment{python}[1][]
{
	\pythonstyle
	\lstset{#1}
}
{}

\begin{document}

\date{}

\title{\Large \bf \system: Taming Hyper-parameter Optimization of Deep Learning
              with Stage Trees}

\author{
{\rm Ahnjae Shin, Do Yoon Kim, Joo Seong Jeong, Byung-Gon Chun}\\
Seoul National University\\
}

\maketitle

\begin{abstract}

\Hpo is crucial for pushing the accuracy of a deep learning model to its limits.
A \hpo job, referred to as a study, involves numerous \trials of training a model using different training knobs, and therefore is very computation-heavy, typically taking hours and days to finish.

We observe that \trials issued from \hpoas often share common \hp sequence prefixes.
Based on this observation, we propose \system, a \hpo system that removes redundancy in the training process to reduce the overall amount of computation significantly.
Instead of executing each trial independently as in existing \hpo systems, \system breaks down the \hp sequences into stages and merges common stages to form a tree of stages (called a stage-tree), then executes a stage once per tree on a distributed GPU server environment.
\system is applicable to not only single \studies, but multi-study scenarios as well, where multiple \studies of the same model and search space can be formulated as trees of stages.
Evaluations show that \system's stage-based execution strategy outperforms trial-based methods such as Ray Tune for several models and \hpoas, reducing GPU-hours and end-to-end training time significantly.

\end{abstract}

\section{Introduction} \label{sec:intro}

Deep learning (DL) models have made great leaps in various areas including image classification~\cite{resnet, cifar, imagenet}, object detection~\cite{yolo}, and speech recognition~\cite{deepspeech, deepspeech2}.
However, such benefits come at a cost; training DL models requires heavy datasets and long computations which may take up to a week~\cite{gnmt} even on a hundred of GPUs~\cite{gnmt}.
This cost becomes more significant when we take \hpo into account.
Since \hps can have great impact on the quality of the trained models, investigating the \hp \searchspace often requires hundreds to thousands of trainings with different \hp settings~\cite{massively}.
Consequently, naively running \hpo requires an exceedingly large number of GPUs, and it is crucial to explore the \hp \searchspace as efficiently as possible.

In this paper, we aim at building a system optimized for running a (possibly multiple) \hpo job, which trains and evaluates the target DL model multiple times, each with a different configuration.
Since each training sub-procedure is identified by its unique configuration, i.e., sampled \hps from a given \searchspace, it is natural to develop a system that can run and manage multiple trainings of the target DL model, especially on a GPU cluster.
Prior works on systems for \hpo attempt to a) efficiently schedule training sub-procedures by considering resource utilization or fairness~\cite{gandiva,themis}, b) provide new abstractions and programming interfaces for productivity of developers~\cite{raytune, optuna, vizier}, c) optimize resource allocation of sub-procedures according to model performance~\cite{asha, hyperdrive, hypersched} or d) design easy to use tuning systems which require minimal coding~\cite{chopt, nni}.
Unlike these works, we explore untapped opportunities to optimize the resource usage of \hpo jobs in terms of the amount of computation. 

Our key observation is that training modern DL models often requires changing {\hp} values in the midst of training to reach state-of-the-art accuracy, as they target minimizing high-dimensional, non-convex loss functions.
Hence, a \hp configuration can be regarded as a \textit{sequence} of values.
Examples include learning rate~\cite{resnet, batchnorm, cycliclr, super, 1hour, adam, adadelta, rmsprop, hypergradient}, drop-out ratio~\cite{elu}, optimization algorithm~\cite{gnmt}, momentum~\cite{yellofin}, batch size~\cite{dont-decay-lr}, image augmentation parameters~\cite{pba}, training image input size~\cite{progan}, and input sequence length~\cite{bert}.

\begin{figure}[t]
  \centering
  \includegraphics[width=0.99\columnwidth]{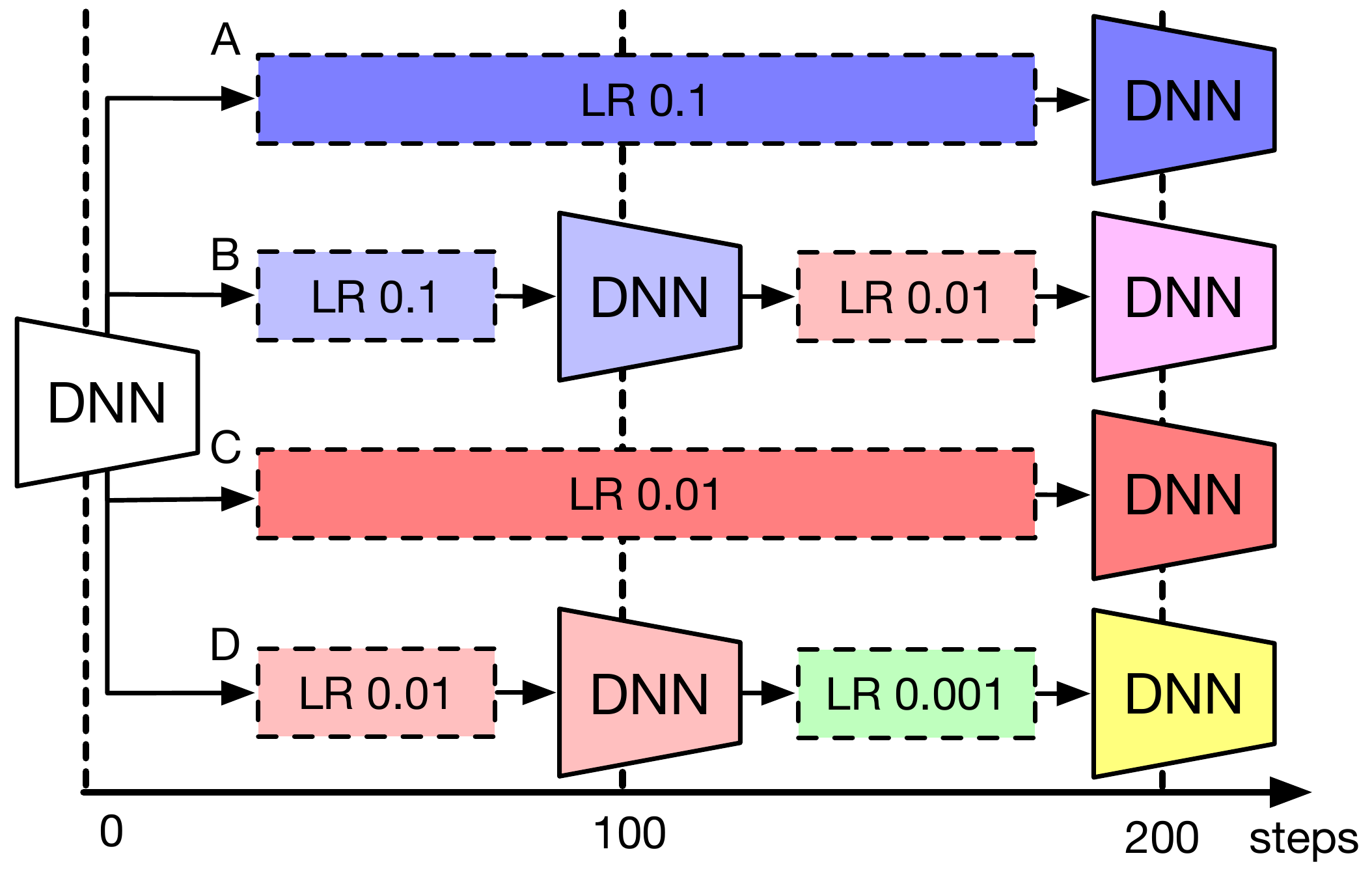}
  \caption{A \hpo \study consisting of four \trials.
           A single \hp, learning rate (LR), is being explored within the \searchspace $\{0.1, 0.01, 0.001\}$.}
  \label{fig:single-study-no-sharing}
\end{figure}

We find that existing approaches for \hpo systems~\cite{raytune, vizier, mltuner, chopt} overlooked this important characteristic of sequential \hps, always treating \hps as single values.
These approaches simply execute multiple training sub-procedures separately, without exploiting the fact that there exist redundant computations between the sub-procedures.
Figure~\ref{fig:single-study-no-sharing} shows a \hpo job with four sub-procedures, each with different learning rate sequence configurations.
A step indicates training with one batch of data.
The first 100 training steps for sub-procedures $A$ and $B$ can be shared, as they are operating on the same learning rate value, $0.1$.
Similarly, $C$ and $D$ also have a common prefix for learning rate $0.01$.
Instead of executing such common prefixes independently, it is possible to execute them only once and share them across sub-procedures to avoid redundant computation and reduce the amount of resources (GPU-hours) used.

To this end, we present \system, a \hpo system that finds redundant computations in \hpo jobs and reuses the results of duplicate workloads.
\system merges \hp sequence configurations in the shape of a tree, called \textit{\tree}, so that all non-leaf nodes represent redundant computations that can be shared.
\Trees also provide the benefit of simplifying the scheduling of \hpo jobs, as each node in the \tree serves as a scheduling unit.
Internally, \system uses an additional data format, a \textit{\plan}, to handle the dynamics of \hpo jobs and manage various states.

We evaluated \system with popular DL models (ResNet56, MobileNetV2, and BERT-Base) and well-known \hpoas (SHA, ASHA, grid search) on a 40-GPU AWS EC2 cluster. Our evaluations show that \system outperforms Ray Tune, a state-of-the-art \hpo system, reducing the end-to-end training time and GPU-hours of a single job up to 2.76x and 4.81x, respectively.
For multi-job scenarios, \system can share redundant computations across jobs and reduce the end-to-end training time and GPU-hours by up to 3.53x and 6.77x, respectively.

The rest of the paper is organized as follows.
Section~\ref{sec:motiv} introduces \hpo and motivates our work.
Section~\ref{sec:stagetree} proposes core representations, \tree and \plan, for identifying and reusing redundant computations.
Section~\ref{sec:system} describes the \system design, and Section~\ref{sec:impl} elucidates implementation details.
Section~\ref{sec:eval} presents evaluation results, Section~\ref{sec:related} explains related work, and Section~\ref{sec:conc} concludes.

\section{Background and Motivation} \label{sec:motiv}

\subsection{\HPO}

\Hpo refers to the act of training multiple instances of a machine learning model with slightly differing training knobs, such as learning rate and batch size. 
We use the term \textit{\study} to refer to a single optimization run of a model over a certain search space of parameters. 
Each sub-procedure of a \study that is associated with a set of parameters sampled from the given search space is called a \textit{\trial}. 

\Hpo is crucial in training deep learning models for high model quality. 
The model quality of \trials with different \hp values may differ significantly, even if settings other than the \hps such as the model architecture and input data are kept the same across all \trials.

There are many types of \hps as well as many possible values for each \hp.
The search space is often very large, and the number of trials is usually in the hundreds and even thousands~\cite{asha, themis, survey}.
As each trial takes up a considerable amount of time, blindly running the trials one after another is impractical for moderately sized cluster environments. 
Many \hpoas are being used throughout the community for quickly finding the trials that yield the best models (in terms of model quality) without executing all trials to completion~\cite{sha, asha, hyperdrive, pbt}.
Meanwhile, various \hpo systems have been proposed to efficiently schedule such trials with average job completion time and inter-user fairness in mind~\cite{gandiva, tiresias, gandiva-fair}.

\textbf{\Hp sequences.}
Many researchers have recently expanded their \hp \searchspaces so that a \hp can change values after a certain amount of steps according to some sequence, rather than being kept as a constant value during the whole \trial.
While the learning rate \hp~\cite{resnet, batchnorm, cycliclr, super, 1hour, adam, adadelta, rmsprop, hypergradient} has always been tuned as a \hp sequence, recent works have also applied this scheme to other \hps as well, such as batch size~\cite{dont-decay-lr}, drop-out ratio~\cite{elu}, optimizer~\cite{gnmt}, momentum~\cite{yellofin}, image augmentation parameters~\cite{pba}, training image input size~\cite{progan}, input sequence length~\cite{bert}, and network architecture parameters~\cite{progan}.
As training modern DL models involves minimizing high-dimensional, non-convex loss functions, we predict this trend of \hp sequences to become even more popular throughout the community.

\begin{figure}[t]
  \centering
  \includegraphics[width=1.0\columnwidth]{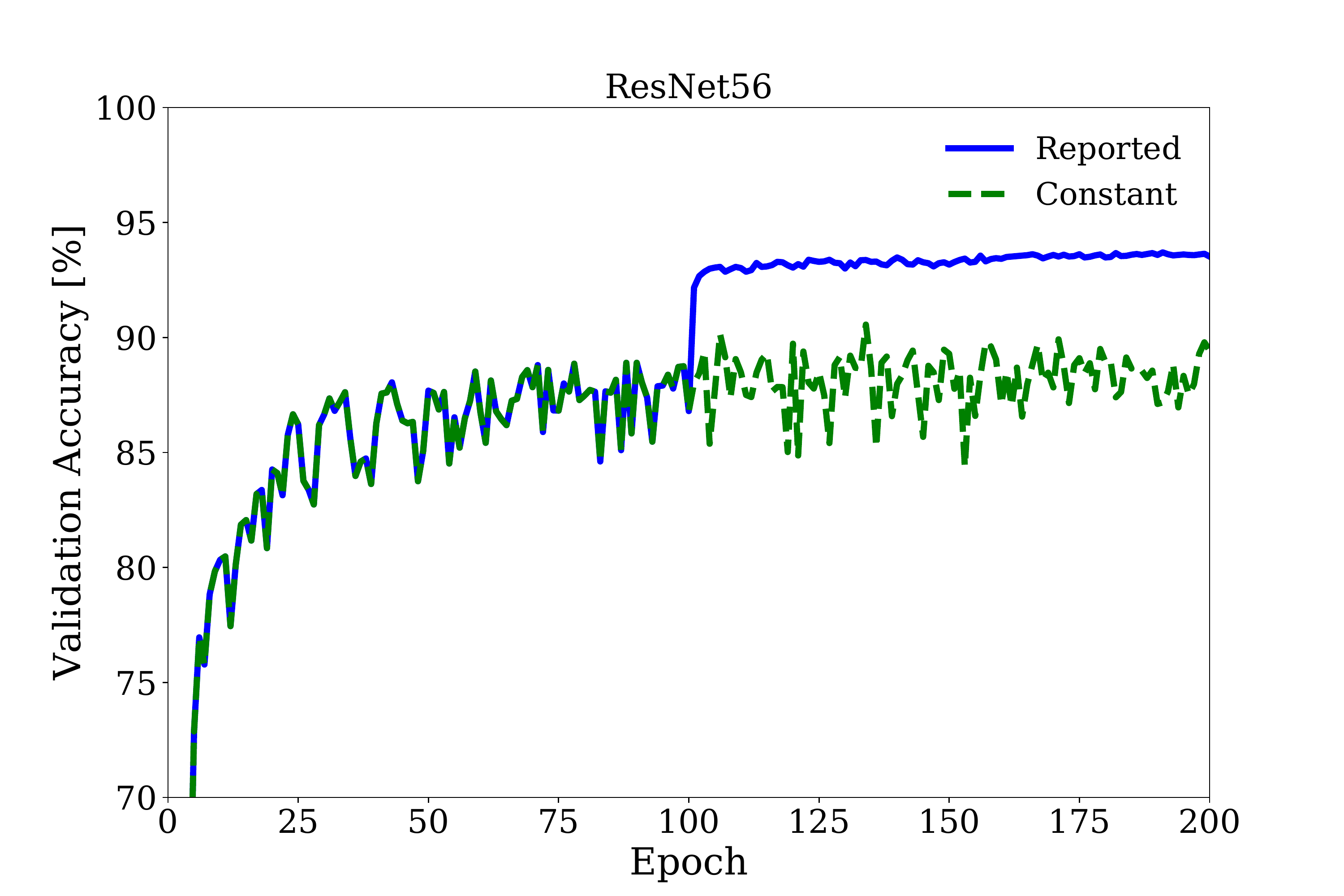}
  \caption{
               Validation accuracy curves for training ResNet56 on CIFAR-10 with various learning rate and batch size settings.
               All other \hps are kept the same as reported in the original ResNet paper~\cite{resnet}.
               \Trial A (green) keeps constant learning rate (0.1) and batch size (128) for the whole \trial, resulting in the lowest accuracy.
               \Trial B (blue) decays the learning rate by 0.1 at the 100th and 150th epochs.}
  \label{fig:motiv}
\end{figure}

We conducted a small \study that consists of two \trials to illustrate the benefits of tuning \hps as a sequence.
Figure~\ref{fig:motiv} shows the model validation accuracy while training a ResNet56 model on the CIFAR-10 dataset, using different learning rate configurations.
\Trial A (green) operates on a constant learning rate, while \Trial B (blue) follows a \hp sequence.
Simply by decaying the learning rate twice, \Trial B reached validation accuracies higher than \trial A by more than 5 percent.
This simple example demonstrates that defining \hps as sequences instead of constant values has great effects on the quality of the trained model.
Clearly, the whole \hp sequence over the course of training, not just the initial value, is important for model quality.

\subsection{Optimization Opportunities}

When generating a new \hp configuration for a trial, many different algorithms each have their own logic on how to generate specific \hp values.
While na\"ive algorithms simply select all configurations in the \searchspace or select a random subset, algorithms such as Bayesian optimization~\cite{bayesian} deliberately sample the next most promising \hps based on the history of \trials, aiming to discover well-performing \trials faster with fewer resources.
Likewise, when manually tuning \hps, a common heuristic to discover a well-performing \trial is to slightly modify a previously attempted \hp sequence that showed good results.
As a result, promising \trials often share common prefixes in their \hps.

\textbf{Sharing computations within a study.}
Figure~\ref{fig:single-study-no-sharing} depicts a hypothetical \hpo \study of \trials that partially share prefixes in their \hp sequences.
Existing \hpo systems treat \trials as black boxes and do not exploit the fact that \trials are actually performing the exact same computation for these prefixes.
By performing such computations only once and reusing the resulting DNN checkpoint multiple times for downstream \trials, we can reduce the amount of GPU-hours required to serve this \study.
As the number of \trials is typically much greater than the number of GPUs available in the training environment, reducing GPU-hours can also lead to reduced end-to-end training time.

\textbf{Sharing computations across studies.}
Reusing computations for common \hp sequence prefixes can be done for multiple \studies, as well.
If a \hpo \study is submitted for a model and a dataset that already had been explored previously by other studies, then we can identify common \hp sequence prefixes among the \studies and reuse past DNN checkpoints to skip redundant computations for the currently submitted \study.

\section{Stage Tree \& Search Plan} \label{sec:stagetree}

Based on our observations in the previous section, we propose two representations for identifying and sharing common computations across \trials and \studies.
We first show a \textit{\tree}, a rearrangement of \trials in the form of a tree that puts common computations at the root and intermediate nodes.
Next, we present a \textit{\plan}, a data layout we use in our system to back \trees.

\subsection{Stage Tree}
\begin{figure}[t]
  \centering
  \includegraphics[width=0.99\columnwidth]{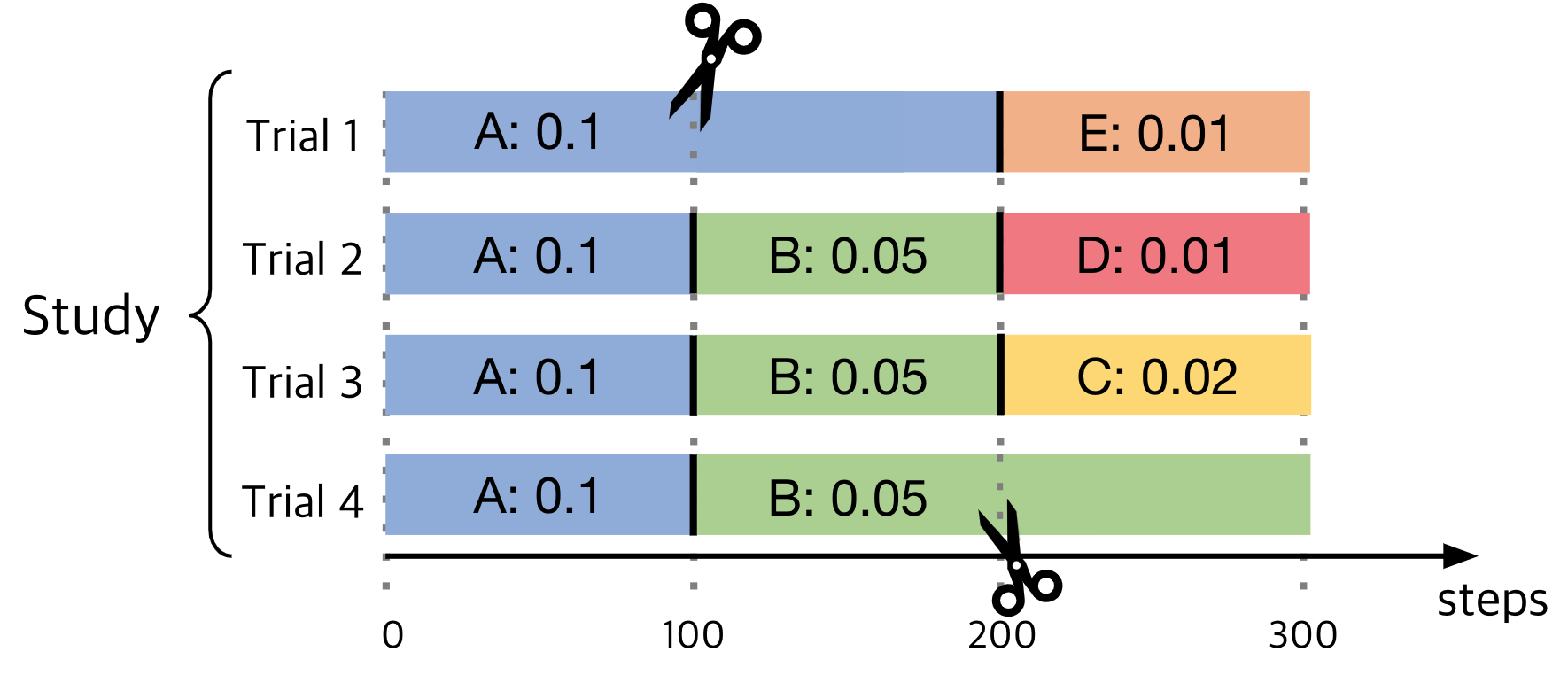}
  \caption{A \study of \trials that share common computations. Each stage is labeled with an id ($A$-$E$) and its \hp value (learning rate $\{0.1, 0.05, 0.02, 0.01\}$). A \stage can be split into shorter \stages to match the length of a \stage from another \study that shares the same \hp value.}
  \label{fig:pic1}
\end{figure}

The trend of employing \hp sequences in \hpo motivates us to divide a \trial into several \textit{\stages}, based on the sequences themselves.
Consider the \study in Figure~\ref{fig:pic1}, consisting of four \trials.
The target \hp being adjusted is the learning rate, and is sampled from a search space of $\{0.1, 0.05, 0.02, 0.01\}$.
\Trial 1 uses a learning rate of $0.1$ for 200 steps, and reduces the value to $0.01$ for the next 100 steps.
In other words, \trial 1 starts with a \stage of learning rate $0.1$ for 200 steps, followed by a \stage of learning rate $0.01$ for 100 steps.
Similarly, \trial 2 is made up of three \stages with 100 steps each.
From now on, we use the term \stage to denote a certain interval of a \trial.
Note that a \stage does not necessarily have to have a constant \hp value;
we could have a \stage that utilizes \hp sequences, such as linear or exponential learning rates.
We follow the convention of dividing \hp sequences to set \stage boundaries as well, such as piecewise linear functions and sequential combinations of functions.

\begin{figure}[t]
  \centering
  \includegraphics[width=0.8\columnwidth]{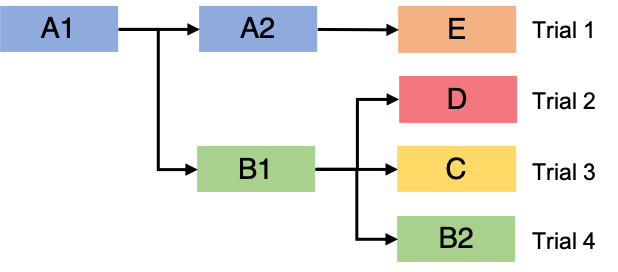}
  \caption{A \tree formed from the trials of Figure~\ref{fig:pic1}. \Stage $A1$ can be executed once to serve all four trials, while \stage $B1$ can be shared by three trials.}
  \label{fig:pic2}
\end{figure}

Dividing \trials into \stages reveals that \trials 2, 3, and 4 actually share the same initial stage (learning rate $0.1$ for 100 steps).
Moreover, the first \stage of \trial 1 can be split into two smaller \stages so that
\trial 1 shares the initial stage as well.
By merging common \stages across \trials, we get the tree-shaped arrangement of \stages in Figure~\ref{fig:pic2}.
In this form, it is evident that \stages $A1$ and $B1$ can be shared by multiple \trials.
We refer to this form as a \textit{\tree}.

The \tree is one of the core representations of a \hp \study in our work, and is mainly used to identify schedulable units when it comes to executing a \hpo study.
Conveniently, a \stage can be considered as a schedulable unit, while edges between \stages express scheduling dependencies.
We show how \trees and \stages are handled to schedule \trials in detail, in Section~\ref{sec:system}.

\begin{figure}[t]
  \centering
  \includegraphics[width=0.99\columnwidth]{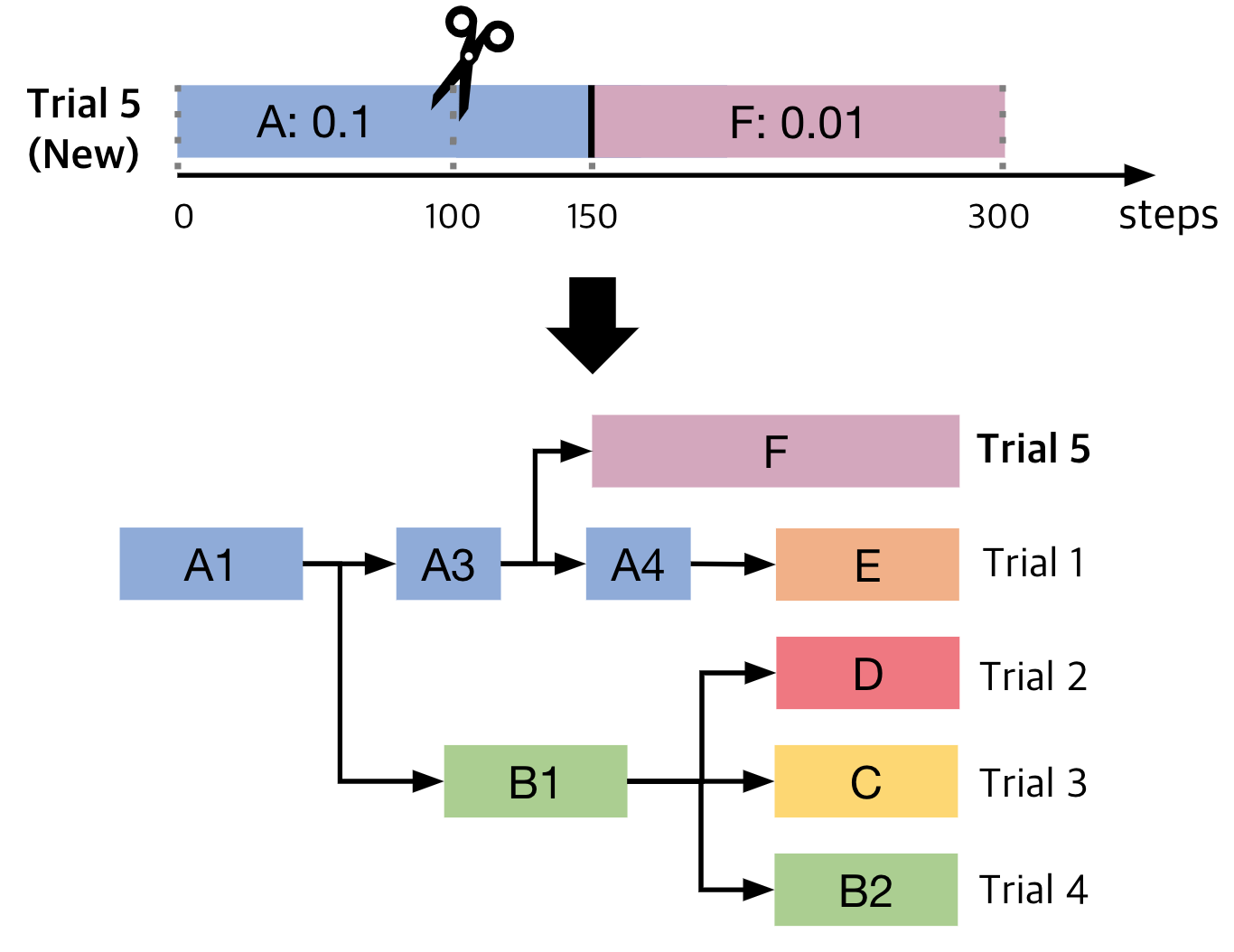}
  \caption{An illustration of a \tree transformation when a new \trial is added to the \tree in Figure~\ref{fig:pic2}. Both the first stage in \trial 5 and \stage $A2$ in Figure~\ref{fig:pic2}'s \tree must be split into smaller stages, in order to merge \trial 5 into the \tree. As a result, \trial 5 shares \stages $A1$ and $A3$ with \trial 1.}
  \label{fig:pic3}
\end{figure}

\subsection{\PLan} \label{subsec:plan}

As new trials arrive, new \stages may be added to a \tree, while existing \stages can be split into shorter \stages of smaller step ranges.
\Stages can even be deleted if the given \hpoa decides to kill certain trials.
For instance, assume that a new trial has been submitted to the previous \tree example, as shown in Figure~\ref{fig:pic3}.
\Stage $A$ of the new \trial (\Trial $5$) cannot be merged into \stage $A1$ or \stage $A2$ in Figure~\ref{fig:pic2}, because neither of them has a matching step range (steps $0$-$150$).
Instead, \stage $A2$ needs to be divided into \stages $A3$ (steps $100$-$150$) and $A4$ (steps $150$-$200$), and then the new \trial's last \stage, $F$, is appended to $A3$.
All \stages that came after $A2$ in the original \tree are modified to follow $A4$ in the new \tree.

Although \tree transformations allow us to map out the current state of a \hpo \study, such dynamics make the implementation of a \tree based system rather complicated for various reasons.
First, from a scheduling perspective, managing execution states of \stages is difficult because \stages can be split even during execution.
For example, if the \stage split from Figure~\ref{fig:pic2} to Figure~\ref{fig:pic3} occurred while \stage $A2$ was still in execution, then it would be unclear how to handle the currently running process.
Second, the ever-changing structure of a \tree makes it difficult to pinpoint the by-products of executing a stage -- namely DNN parameter checkpoints and validation accuracy values -- which must be associated with specific \hp sequences.

To resolve such issues, we introduce another internal representation for a \hpo study, the \textit{\plan}, that is similar to a \tree but does not involve any node removals for a newly submitted \trial.
Nodes contain information of past \trials that passed through those particular nodes, while each edge is annotated with an integer indicating the \stage boundary in the number of steps.

\begin{figure*}[h]
  \centering
  \includegraphics[width=1.0\textwidth]{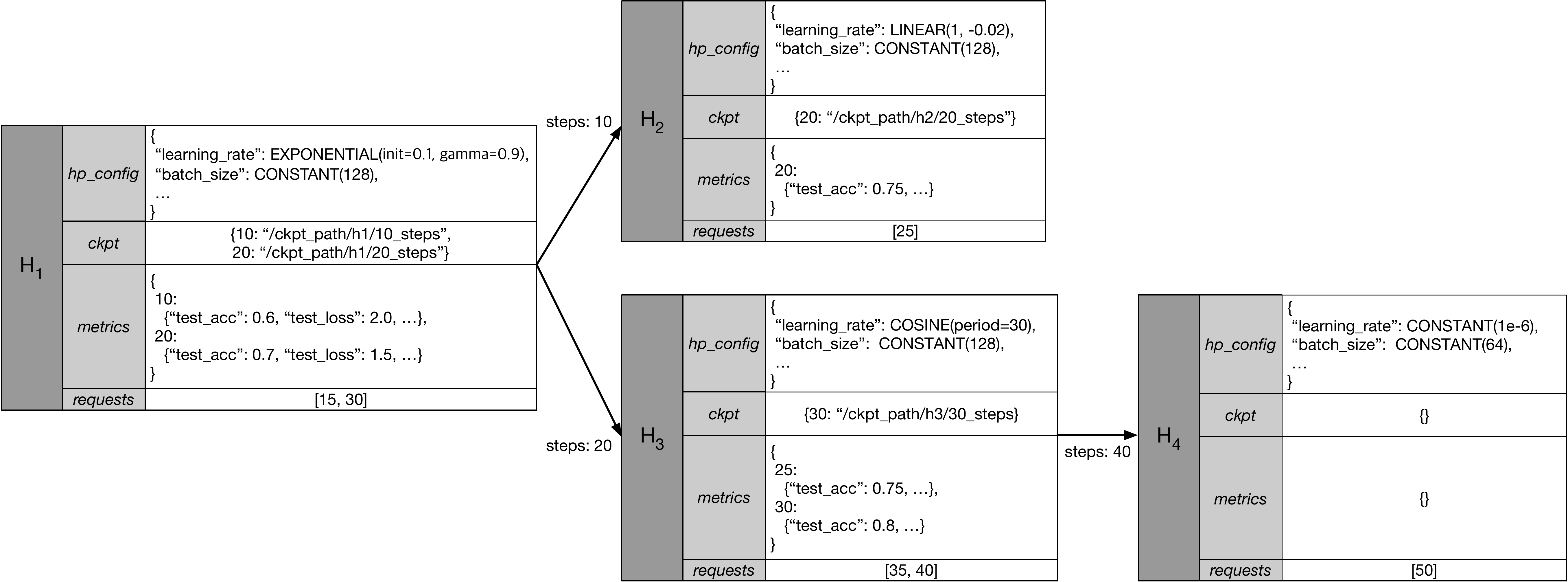}
  \caption{A \plan example of \hp configurations. Each node stores various fields, including \hp value functions for each \hp (\texttt{hp\_config}), checkpoints and intermediate values for later reuse (\texttt{ckpt}, \texttt{metrics}), and a list of integers that mark the current \stage(s) that are waiting to be executed under this configuration (\texttt{requests}). Edges across nodes indicate sequential dependencies, e.g., $H_3$ occurs after training a model for 20 steps under $H_1$, while $H_4$ occurs after training a model for 20 more steps under $H_3$ (a total of 40 preceding steps).}
  \label{fig:plan}
\end{figure*}

\begin{figure}[t]
  \centering
  \includegraphics[width=0.99\columnwidth]{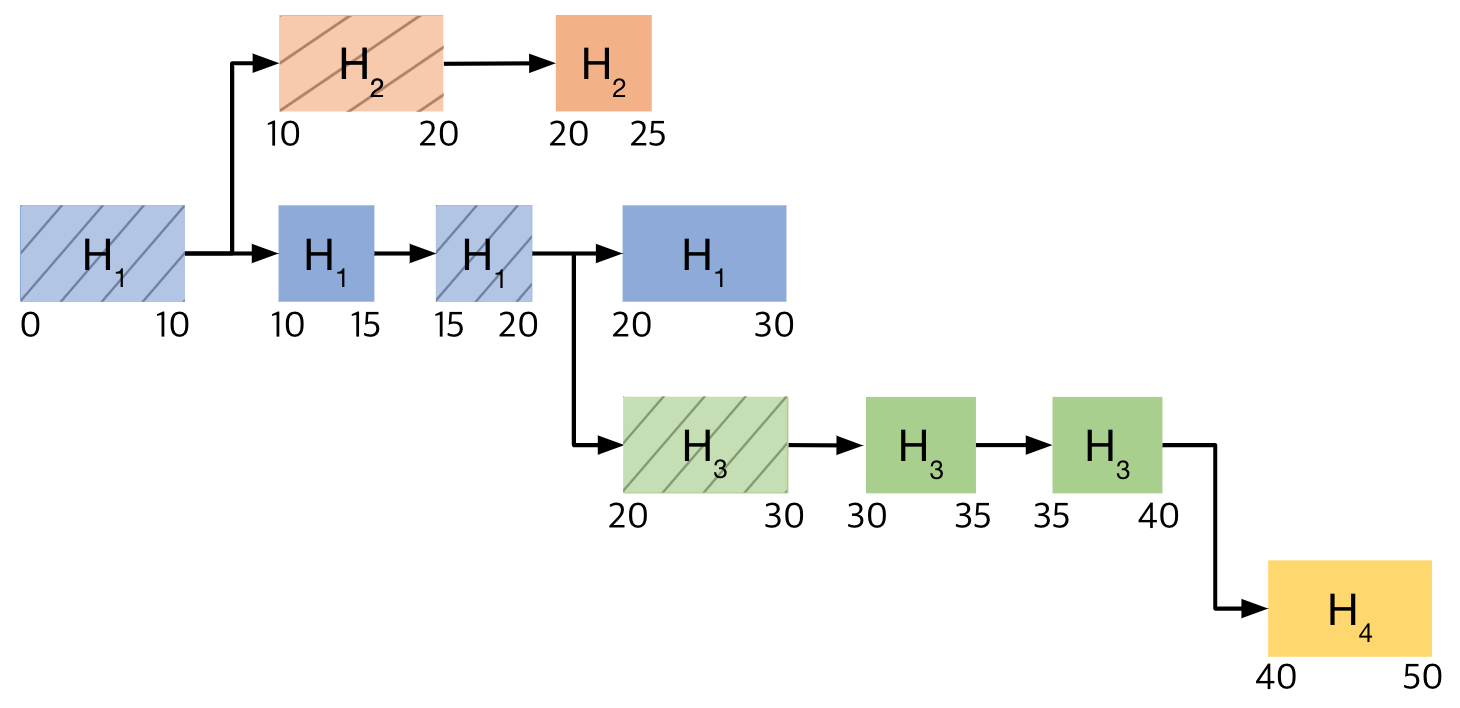}
  \caption{A \tree generated from the \plan in Figure~\ref{fig:plan}.
  The numbers below each \stage indicate the step to start and stop training.
  Shaded \stages indicate \stages with checkpoints where training can be resumed from.
  }
  \label{fig:pic4}
\end{figure}

An example of a \plan is drawn in Figure~\ref{fig:plan}.
Each node represents a \hp configuration starting from a certain training step.
$H_1$, the root node of this \plan, indicates a configuration of training a freshly initialized model (no parent node) with an exponential learning rate and constant batch size.
Likewise, $H_2$ indicates a configuration of a linear learning rate and constant batch size, starting from a model checkpoint that has been trained with $H_1$ for 10 steps (note the directed edge between $H_1$ and $H_2$).

Unlike \trees, a \plan node is not a scheduling unit.
The existence of a node does not necessarily imply that a \trial, configured by that node, is currently running in the system.
Rather, a node holds various statistics gathered by the system regarding the corresponding \hp configurations, specified by the following fields:

\begin{itemize}
  \item \textit{hp\_config}:
  \Hp configurations for each target \hp, given as functions.
  Values for coefficients and constants are also given, if required (not shown in the figure).
  Widely used functions for \hp values, such as \texttt{CONSTANT}, \texttt{EXPONENTIAL}, \texttt{COSINE}, and \texttt{STEP}, are allowed.

  \item \textit{ckpt}:
  A dictionary of file paths for checkpoints that were trained under this configuration.
  Dictionary keys are used to indicate the number of training steps.

  \item \textit{metrics}:
  Intermediate values for evaluating the quality of the model, like test/validation accuracy and loss.

  \item \textit{requests}:
  A list of integers representing requests to train and return the metrics of submitted \stages.
  Each integer indicates the total number of training steps required per request.
  A request consists of one or more \stages with sequential dependencies, which are expressed by an integer in a node's requests field and the edges connecting from preceding nodes.
  Note that one request may map to one or more \trials if they are merged into the same \stage(s).
  For example in Figure~\ref{fig:plan}, 15 in $H_1$'s requests field indicates that one or more \trials require training with $H_1$'s \hp configuration for 15 steps.
  To interpret 35 in $H_3$'s requests field in a similar manner, we must first follow the edge from the preceding \stage ($H_1$), indicating that the request requires training for 20 steps with $H_1$'s \hp configuration.
  Then, the request requires training for an additional 15 steps with $H_3$'s \hp configuration, for a total of 35 steps.
  Note that since there already is a checkpoint for $H_1$ at 20 steps, training can be resumed from this checkpoint.

\end{itemize}

We also have a few additional fields for implementation reasons, such as a reference count value and other runtime metadata.
We will explain these further in later sections.

When a new \trial arrives, it is first compared with the \plan to see if there exists a path from the root node to a leaf node that exactly matches the \trial's \hp sequence.
If not, then a new node is created and added to the \plan.
Otherwise, we next check the \textit{ckpt} and \textit{metrics} fields of the leaf node and immediately return the appropriate results in case no training is needed (e.g., there already is a checkpoint that matches the \trial's \hp sequence).
For the more common case in which we need to train the model, then an entry is added to the \textit{requests} field and a \tree is generated from the \plan, to be handed over to a separate scheduler component.

Going back to the example illustrated in Figure~\ref{fig:pic3} where a new \trial submission requires splitting an existing \stage (A2), \system handles this case by simply adding another item to an existing node's requests field to indicate a new \stage (A3) with the same \hp configuration but a smaller number of steps.
Note that if A2 was executed before this change and a checkpoint for A3 was not made, computation for A3 may be repeated again, later when a new \tree is generated from this updated \plan.

\begin{algorithm}[t]
	\caption{Build Stage Tree}
	\label{alg:plan-to-tree}
	\begin{algorithmic}[1]
		\Function{BuildStageTree}{$x$}
		\State Initialize lookup table $L$
		\For{each not scheduled request~$r$}
		\State FindLatestCheckpoint(r, $L$) // r is (hp\_config, step)
		\EndFor
		
		\State Initialize a tree $S$
		\For{each key, value in $L$}
		\State // each key and value is a request object
		\State // create a stage that load checkpoint at value and train until key.
		\EndFor
		\State connect consecutive stages
		\State \Return $S$
		\EndFunction
		
		\Function{FindLatestCheckpoint}{$r$, $L$}
		\If{r.hp\_config is running}
		\State \Return $L$.put($r$, null)
		\EndIf
		\If{$r$ has no parent configuration || $r$ $\in$ $L$}
		\State \Return
		\EndIf
		\For{s $\in$ \{step, step - 1,  ..., hp\_config.start\}}
		\If{Checkpoint $C$ exists at (hp\_config, s)}
		\State\Return $L$.put($r$, (hp\_config, s))
		\EndIf
		\EndFor
		\State $r_p$ = (hp\_config.parent, hp\_config.start)
		\State $c$ = FindLatestCheckpoint($r_p$, $L$)
		\State $L$.put($r$, $r_p$)
		\State \Return 
		\EndFunction
	\end{algorithmic}
\end{algorithm}

\textbf{Going from \plans to \trees.}
While \plans are effective for managing the current status and history of a \hp \study, \stages are more straight-forward as a scheduling unit for a system scheduler component to interact with.
Thus, we use \plans as the basic format for holding internal data, but ultimately generate \trees when a scheduling decision needs to be made.
The generated \trees are transient representations, used solely for creating scheduling units (\stages), and are not kept in the system like \plans.

Algorithm~\ref{alg:plan-to-tree} describes the process of generating a \tree from a \plan. We first think of a helper function \textit{FindLatestCheckpoint}. This function gets a request object and a lookup table as input. The request object is a tuple of hyper-parameter configuration and step. \textit{FindLatestCheckpoint} finds the closest checkpoint to the request and stores it to $L$. If the closest checkpoint belongs to one of its ancestors, the function recursively calls with its parent as an input in line 27. Also, it adds its parent to the lookup table in line 28. Therefore, after the lookup table is completed in line 5, the algorithm goes over each entry in the lookup table and starts building the stage tree.
Note that the lookup table is also used as a memoization mechanism as in line 18.

Figure~\ref{fig:pic4} illustrates the \tree generated from the \plan in Figure~\ref{fig:plan}.
A \stage will be executed by resuming from the nearest available checkpoint, where available checkpoints are marked as shaded areas in the figure.
For example, the stage denoted by $H_2$ with steps 20 to 25 in the figure will be trained after resuming from $H_2$'s checkpoint at 20 steps, which can be seen in the \textit{ckpt} field of $H_2$ in Figure~\ref{fig:plan}.

\section{\system System Design} \label{sec:system}

In this section, we introduce \system, a \hpo system that incorporates \trees and \plans to run multiple \studies while automatically reusing computation for sharable \stages within a study and across studies.

\begin{figure}[t]
    \centering
    \includegraphics[width=0.8\columnwidth]{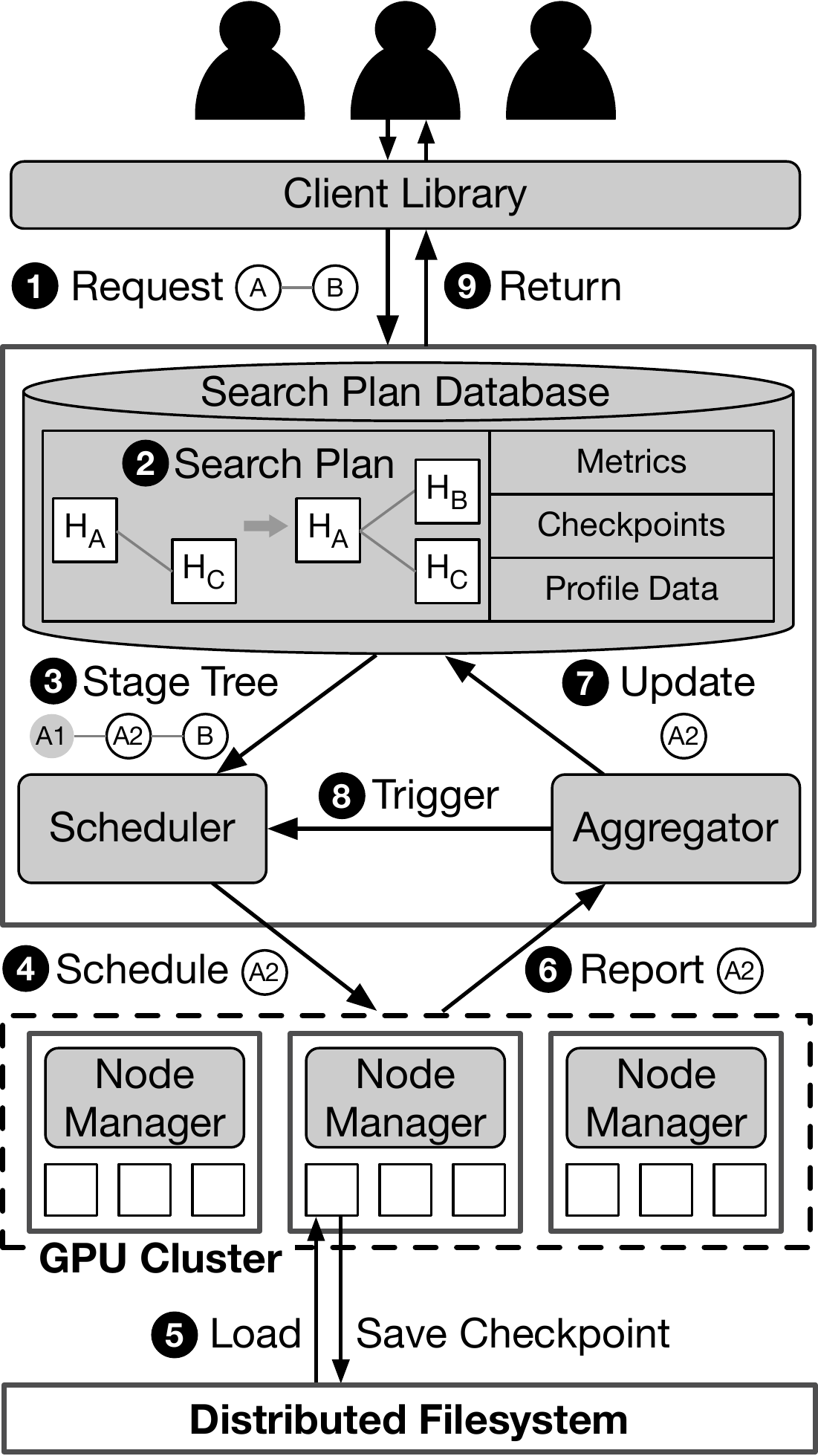}
    \caption{
      \system system architecture.
    }
    \label{fig:system}
\end{figure}

\subsection{Overview}
Executing a \trial in \system is initiated by the user submitting the \trial to \system via the \textit{client library}, a thin interface for constructing \trial requests in the appropriate format.
A \trial request is defined as a pair of a \hp sequence configuration and the number of training steps (Figure~\ref{fig:system} \circled{1}).
Once a request arrives at the system, the \hp sequence configuration is immediately compared with the corresponding \plan in the \textit{\plan database}, and the \plan is adjusted accordingly (\circled{2}).
In case metrics and checkpoints that satisfy the request's criteria are already present, then a response is returned immediately to the user.
Otherwise, the \textit{scheduler} is notified to run new \stages.

The scheduler decides which \stages to run by examining the \tree generated from the current \plan (\circled{3}).
\Stages are given to GPU workers for execution (\circled{4}), and the workers start computation by loading checkpoints from the distributed filesystem (\circled{5}).
Workers periodically report evaluation metrics to the \textit{aggregator} (\circled{6}).
Each server has a node manager to gather metrics locally before passing them to the aggregator, for reducing inter-server data traffic.
The aggregator, upon receiving a set of metrics, updates the \plan (\circled{7}) and also pings the scheduler if a new checkpoint has been added (\circled{8}).
After repeating the scheduler-aggregator cycle multiple times, the final \stage for a \trial request will eventually terminate, and the metrics are sent back to the client (\circled{9}).

\subsection{\PLan Database}

\system stores all \plans that are currently being served in the \plan database.
When a new \trial is added, \system updates the \plan as described in Section~\ref{subsec:plan}.
The various field entries in any node of the \plan, including checkpoints, metrics, and runtime profile data, can also be updated by the aggregator component.

\subsection{Scheduler}

\system schedules computation on GPUs with \stages as the basic scheduling unit.
Since the number of \stages that can run concurrently at a given moment usually exceeds the number of available GPUs in the cluster, \system utilizes a scheduler component to determine the \stages to be run.

The scheduler takes a \tree generated from the current \plan as input, and schedules \stages on GPU workers.
A simple scheduling method would be to do a breadth-first traversal through the \tree and schedule each \stage one by one, until all GPU workers have been assigned a \stage.
However, we have found that this method incurs significant transition overhead for workers because the scheduling granularity (\stage) is too small.
Instead, the scheduler computes the critical path of a given tree and schedules the whole sequence of \stages in the path on a worker.
With multiple workers, the scheduler repeats finding the next critical path among unscheduled \stages in the \tree and scheduling the sequence of \stages on an idle worker.
The larger scheduling granularity (batch of \stages) not only improves locality by avoiding overheads such as checkpoint save/loading, but also allows us to prioritize minimizing the end-to-end training time of a \hpo \study.
The critical path of a \tree is the path (from root to leaf) that has the longest estimated execution time;
the execution time of an individual \stage is estimated by multiplying the number of steps of that \stage by the execution time per step (profiled beforehand and stored in the \plan database).

Note that the scheduler does not store any information regarding the execution states of \stages.
Since \stages can be split and even removed during execution (as mentioned in Section~\ref{sec:stagetree}), setting \stage states to handle the execution of a \study involves complex state management measures.
Instead, the scheduler operates in a stateless manner, relying entirely on the \plan to identify the \stages that need to be run and the \stages that have already run.
In other words, after scheduling \stages from a \tree, the scheduler simply 
releases the \tree.
When the scheduler is triggered later to schedule more \stages, the scheduler takes a new \tree freshly generated from the latest \plan, and repeats finding and scheduling the next critical path of unscheduled \stages in the \tree on an idle worker. 

\section{Implementation} \label{sec:impl}

We have implemented \system in Python, using various libraries.
The system utilizes Python's concurrent programming library, \texttt{asyncio}, to manage coroutines.
Communication between the main \system process and node managers is done via the pub/sub interface provided by Apache Kafka 2.4.1, together with Apache ZooKeeper 3.4.13.
MySQL 8.0 is used to store system states in the \plan database.
Kafka, ZooKeeper, and MySQL are all run in Docker containers.
Additionally, we use GlusterFS 6.9 as the distributed file system for saving and sharing checkpoints between nodes.
Our current implementation of \system utilizes the deep learning framework PyTorch 1.5.0 to train DNN models, though \system's design is not tied to any specific framework.

\subsection{Data Pipeline}

We implemented a custom data pipeline for PyTorch that is compatible with \stages.
Two major updates were done.
First, we modified the checkpoint mechanism of PyTorch's default data pipeline to include the current permutation of the dataset as a part of the checkpoint.
This way, the data pipeline is able to save its current position in the dataset when a \stage terminates, and later resume from the same position for the next \stage.
Second, we added a feature to change the batch size of the data pipeline.
When the batch size is changed, the data pipeline will flush every preprocessed batch from the queue, and relaunch the background threads so that they produce the correct batch samples.

\subsection{Client Library}

\begin{figure}[t]
	\begin{lstlisting}[language=Python,breaklines=true,basicstyle=\ttfamily,tabsize=2]
	class MyTrainer(Trainer):
		...
		def setup(self, hp): 
		# hp is a dictionary of updated values
			if "lr" in hp:
				for group in self.optimizer.param_groups:
					group["lr"] = hp["lr"]
			if "bs" in hp:
				if self.train_loader:
					del self.train_loader
				self.train_loader = DataLoader(
					self.train_dataset,
					batch_size=hp["bs"]
				)
		...
	\end{lstlisting}
	\caption{An example that updates the learning rate (lr) and batch size (bs) in the custom \texttt{Trainer} that the user should override. \system passes into \texttt{setup} the values of sequential \hps that should be updated.}
	\label{fig:api-trainer}
\end{figure}

\begin{figure}[t]
	\begin{lstlisting}[language=Python,breaklines=true,basicstyle=\ttfamily,tabsize=2]
	def get_search_space():
		hp = {
			"lr": [
				Constant(0.1),
				Exponential(0.1, 0.95)
			],
			"bs": [
				Constant(128),
				MultiStep(128, [40], 2)
			]
		}
		return GridSearchSpace(hp)
	\end{lstlisting}
	\caption{Defining a search space consisting of learning rate (lr) and batch size (bs) sequences in Python using the function definitions provided by \system. Two different sequences were defined for each \hp, resulting in four \trials.}
	\label{fig:api-searchspace}
\end{figure}

\begin{figure}[t]
	\begin{lstlisting}[language=Python,breaklines=true,basicstyle=\ttfamily,tabsize=2]
	hp_set = ["lr", "bs"]
	study = Study(remote_url).create(
		dataset, command, ckpt_path, hp_set
	)
	
	schedule = Schedule.from_milestones(
		(5, 8), (10, 4)
	)
	tuner = EarlyStopTuner(
		schedule, search_space,
		metric.ExtractSingleNumber(
			"test_acc"
		)
	)
	tuner(study)
	
	# Users can tune a study multiple times on different tuners
	tuner2(study)
	
	# Users can directly evaluate a certain trial on a specified step
	study.eval(hp_config, step)
	\end{lstlisting}
	\caption{Running a study with an example tuner that trains 8 \trials for 5 logical training iterations, early-stops 4 \trials, and trains the remaining 4 \trials up to 10 logical iterations. The killing decision is made based on the test accuracy as specified in the last argument to \texttt{EarlyStopTuner}.}
	\label{fig:api-tuner}
\end{figure}

To run a \study in \system, users must first decide the model and dataset they want to use in the \study, the types and values of \hps to tune, and the tuning algorithm to use.
The training logic, which describes all things needed for training a model such as setting the values of each \hp, is defined by overriding the base \texttt{Trainer} class \system provides.
The values of each \hp used in the \texttt{Trainer} will be drawn from the search space defined in Python by the user. 
The tuning algorithm specifies how to spawn, pause, or terminate \trials that compose the \study.
Users may implement their own strategies, or simply choose from the tuners we provide.
We will now take a closer look at each step a user must take to run a \study in \system.

First, users should implement the training logic by overriding the base \texttt{Trainer} class \system provides.
Users should write functions for initializing training (e.g. defining the model or loading the dataset), training for one logical iteration (which may consist of multiple steps), evaluating the model trained so far and returning the metrics, saving, and loading checkpoints.
One logical training iteration, executed by one call to the \texttt{Trainer}'s \texttt{train} function, should be long enough to avoid overheads, but short enough to regularly report progress. Often, a logical training iteration is set as one pass through the dataset.
Whenever a \hp value is initialized or updated within a \stage, \system calls the \texttt{Trainer}'s \texttt{setup} function with a dictionary containing updated values. Then, using theses values in \texttt{setup}, the user should make according changes to the appropriate attributes of the \texttt{Trainer}.
Figure~\ref{fig:api-trainer} illustrates a \texttt{setup} example.

Then, the user should define the search space they wish to explore using \system's implementation of well-known functions.
Figure~\ref{fig:api-searchspace} displays a simple example that creates a search space over two types of \hps to use with the \texttt{MyTrainer} class defined previously in Figure~\ref{fig:api-trainer}.
Unlike in existing frameworks, users can directly express \hps in the search space as sequences, without having to embed the sequences as part of the training logic.
Notice the matching keys between the search space and the \texttt{hp} dictionary passed into \texttt{MyTrainer}'s \texttt{setup}.
\Trials are sampled from this search space as a grid here, resulting in a total of four \trials, but users who wish to implement conditional \hp spaces can optionally pass in a function to \texttt{GridSearchSpace} to filter out certain \trials.

The last step is to create a \study and a tuner.
A \study is defined by specifying the dataset, the command to run a \trial, the checkpoint path, and the \hp set.
The \hp set contains the types of \hps that are tuned in the study.
For tuners, we provide several \hpoas such as Successive Halving (SHA)~\cite{sha}, Hyperband~\cite{hyperband}, Asynchronous Successive Halving (ASHA)~\cite{asha}, median-stopping~\cite{vizier}, and PBT~\cite{pbt} in the client library.
Figure~\ref{fig:api-tuner} illustrates how to create a study with a search space containing two types of \hps, and tune the study with a tuner that early-stops \trials on milestones based on a certain metric.

\system's client library heavily utilizes Python's \texttt{asyncio} library.
Instead of creating a new thread for each request, the library creates coroutines which are handled by the default Python event-loop.
The tuning algorithms provided by \system take advantage of \texttt{asyncio} primitives, such as \textit{wait\_all} (block until all coroutines have finished) and \textit{wait\_any} (block until at least one coroutine has finished), to implement their logic.

Typically, \hpoas submit several requests in parallel.
In such situations, the client library batches the requests to reduce processing overhead at the \plan database.

\section{Evaluation} \label{sec:eval}

We compare \system to Ray Tune \cite{raytune}, a \hpoa framework built on top of Ray \cite{ray}, to present the experimental results that show how a \study can be executed both quickly and efficiently on \system.
We conducted four single study experiments comparing Ray Tune and \system, and two multi-study experiments each with a varying number of studies that run in parallel.

\begin{table*}[ht]
    \centering
    \begin{tabular}{|c|c|c|c|c|c|}
    \hline
    Model       & Dataset   & Tune Algorithm & Algorithm Policy & \# of trials & Merge rate (p) \\ \hline
    ResNet56    & CIFAR-10   & SHA            & reduction=4, min=15, max=120                & 448          & 2.447               \\ \hline
    ResNet56    & CIFAR-10   & ASHA           & reduction=4, min=15, max=120                 & 448          & 2.447               \\ \hline
    MobileNetV2 & CIFAR-10   & Grid search    & max=120        & 240          & 3.144               \\ \hline
    BERT-Base   & SQuAD 2.0 & Grid search    & max=27000                 & 40           & 2.045                \\ \hline
    \end{tabular}
    \caption{Specification of four studies. Each study is specified a model, dataset, tuning algorithm, tuning algorithm policy, and a search space.
    Min and max are number of steps for BERT-Base, and number of epochs otherwise.
    The number of trials and merge rate of each search space is provided.}
    \label{tbl:single-study}
\end{table*}

\paragraph{Environment}

All experiments were conducted on Amazon Web Services. 
Each experiment uses a homogeneous GPU cluster of five p2.8x instances, each with 8 NVIDIA Tesla K80 GPUs. 
A distributed file system using GlusterFS \cite{gluster} is set up on Amazon EBS volumes, with one volume per instance. 
For trials that do not fit in one GPU, we apply synchronous data parallel training. 
All experiment scripts are implemented in PyTorch 1.5.0 \cite{pytorch}. 
In all of our experiments, we measure the end-to-end time, the elapsed time from the start of the experiment to the end, and the GPU-hours, the sum of elapsed time each GPU was held for training.

As an effort for a fair comparison, we have made the following changes. 
First, we have re-implemented the ASHA \cite{asha} algorithm on Ray Tune, as the implementation provided by Ray Tune was different from the original paper.
Second, we altered Ray Tune's Trainer implementation so that it performs its evaluation multiple times in each of its epoch. 
In addition, we match the number of evaluations between Ray Tune and \system.

\paragraph{Merge rate}

As our evaluation results vary on the configuration of the search space, we provide a coefficient $p$ that summarizes the merging capability of our search space.

\begin{align*}
    p = \frac{\text{Total training iterations}}{\text{Unique training iterations}}
\end{align*}

\textit{Total training iterations} is defined as the number of training iterations that are needed to train the entire search space without reusing any computations. 
\textit{Unique training iterations} is defined as the number of training iterations with zero redundant computation.
For example, if there are N identical trials, the merge rate $p$ is $N$.
Note that to bound our iteration counts, the number of iterations for each trial is set as the maximum iterations a trial can be trained.

Similarly, we define a $k$-wise merge rate $q$ defined on $k$ search spaces.

\begin{align*}
    q = \frac{\text{Total training iterations of K studies}}{\text{Unique training iterations across K studies}}
\end{align*}

Merge rate is not the only coefficient that determines the GPU-hour reduction in \system. 
The total GPU-hour is also affected by the difference in the training durations for each trial. 
Also, when applying a \hpoa that early-stops under-performing trials, the actual GPU-hour is affected by the \trial that is early-stopped.
If a trial that shares common computation with many other trials is early-stopped, the overall GPU-hour reduction will decrease.

\paragraph{Search space}

Over four studies, we use a total of seven types of \hps. 
Five \hps (learning rate, batch size, momentum, image cutout\cite{cutout} size, input sequence length\cite{bert}) were tuned as a sequence and two \hps (optimizer, weight decay) were tuned as a single value throughout the trials.
We do not optimize the hyper-parameter sequences directly. 
Instead, we follow the convention of parameterizing the \hp sequence as a well known function, and tune the parameters instead. 
For example, the learning rate search space is composed of the following functions: step-decay, linear, exponential decay after linear warm-up, and other functions. 
The parameters of each function were chosen to build a search space. 
For example, only a few learning rate sequences were generated from the linear function, which varies on the slope and the initial value.
For the learning rate, we sample many different sequences from the commonly used functions implemented in most DL frameworks. 
For other \hps, a constant value or piecewise constant sequences were used.  
Each search space varies on the type of \hps, number of trials, and the sequence values of each \hp. 
We provide an overview of the search space for each of the studies, as well as the statistics, like the number of trials, and the merge rate $p$ or $q$.

\subsection{Single Study}

\begin{figure*}[ht]
    \centering
    \begin{subfigure}[b]{0.49\textwidth}
       	\centering
        \includegraphics[width=1.0\textwidth]{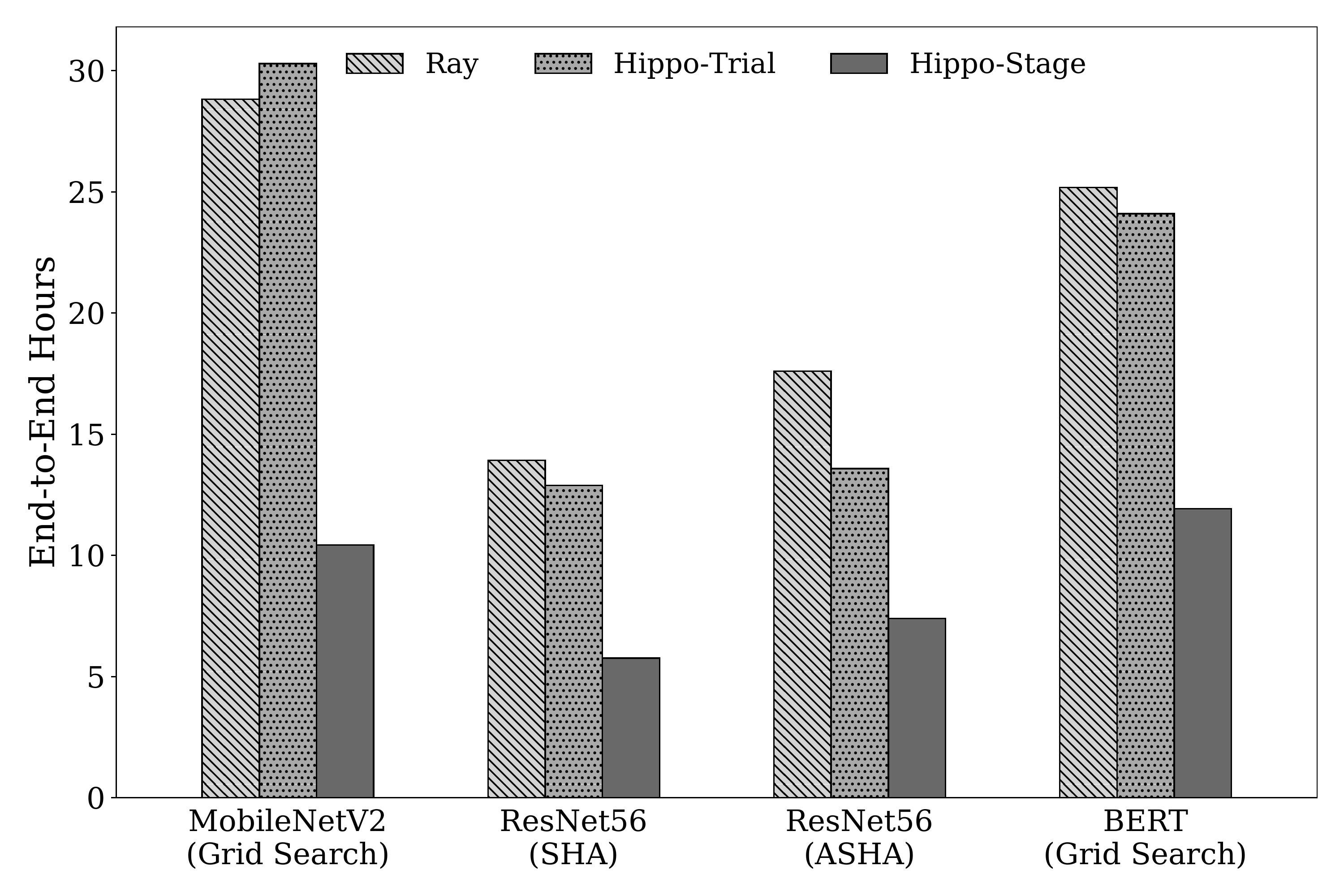}
        \centering
       	\caption{End-to-end time}
       	\label{fig:single-e2e}
    \end{subfigure}
    \hfill
    \begin{subfigure}[b]{0.49\textwidth}
        \includegraphics[width=1.0\textwidth]{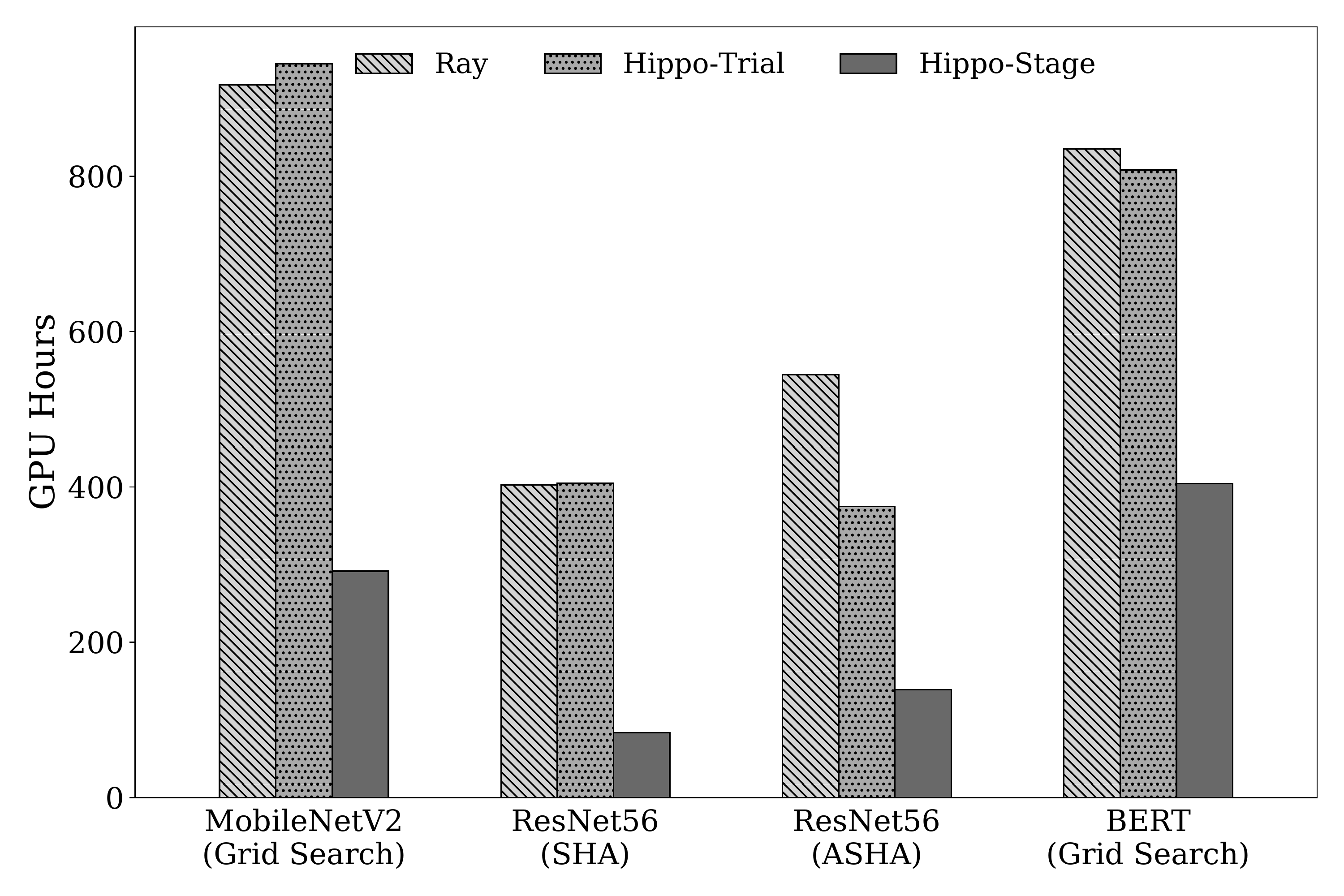}
    	\caption{GPU-hour}
    	\label{fig:single-gpu}
    \end{subfigure}
    \caption{
        Single Study Experiment results
    }
    \label{fig:single}
\end{figure*}

This section compares three different \hpoa systems: Ray Tune, \system, and \system-trial. 
Along with \system, we provide the same evaluation results with \system-trial, an implementation of \system but without merging so that no computation is reused.

\paragraph{Studies}

We compare four different studies across three different \hpoa systems. 
The design of each study is described in Table~\ref{tbl:single-study}.
Three different models, two different datasets, and three different \hpoas are used for the different studies. 
For the ResNet56 and MobileNetV2 models, each trial does not train for more than 120 epochs. 
Only the trial with the highest accuracy is trained for 100 additional epochs. 
The extra training of the best performing trial is accounted to the GPU-hour and the end-to-end time.

\begin{table}[ht]
    \centering
    \begin{tabular}{|c|c|}
    \hline
    \makecell{Hyper\\Parameter}                & Function family                     \\ \hline
    \multirow{5}{*}{learning rate} & \makecell{Initial=0.1, StepLR(gamma=0.1,\\milestones=[90,135])}                         \\ \cline{2-2} 
                                   & \makecell{Warmup(5,0.1), StepLR(gamma=0.1,\\milestones=[90,135])} \\ \cline{2-2} 
                                   & \makecell{Warmup(5,0.1), \\ Exponential(gamma=0.95)} \\ \cline{2-2} 
                                   & \makecell{Warmup(10,0.1), \\ CosineAnnealingWarmRestarts($t_{0}$=20)} \\ \cline{2-2} 
                                   & \makecell{CyclicLR(base\_lr=0.001, max\_lr=0.1,\\step\_size\_up=20)} \\ \hline
    \multirow{2}{*}{batch size}    & 128                            \\ \cline{2-2} 
                                   & values=(128,256), milestones=[70]  \\ \hline
    \multirow{2}{*}{momentum}      & 0.9                            \\ \cline{2-2} 
                                   & values=(0.7,0.8,0.9), milestones=[40,80] \\ \hline
    weight decay                   & 1e-4, 1e-3                           \\ \hline
    optimizer                      & \makecell{Adam, Vanilla SGD, \\ SGD with nonzero momentum} \\ \hline
    \end{tabular}
    \caption{Examples from the search space defined for ResNet56. 5 types of hyper-parameters were tuned.
    Warmup indicates the duration in epochs and target value.}
    \label{tbl:resnet-searchspace}
\end{table}

\begin{table}[ht]
    \centering
    \begin{tabular}{|c|c|}
    \hline
    \makecell{Hyper\\Parameter}                                                                       & Function family                     \\ \hline
    \multirow{5}{*}{learning rate}                                                        & \makecell{Initial=0.1, StepLR(gamma=0.1,\\milestones=[100,150])} \\ \cline{2-2} 
                                                                                          & \makecell{Warmup(10,0.1), StepLR(gamma=0.1,\\milestones=[100,150])}    \\ \cline{2-2} 
                                                                                          & \makecell{Warmup(10,0.1), \\ Exponential(gamma=0.95)}   \\ \cline{2-2} 
                                                                                          & \makecell{Warmup(10,0.1), \\ CosineAnnealingWarmRestarts($t_{0}$=20)} \\ \cline{2-2} 
                                                                                          & \makecell{CyclicLR(base\_lr=0.001, max\_lr=0.1,\\step\_size\_up=20)}\\ \hline
    \multirow{2}{*}{batch size}                                                           & 128                                \\ \cline{2-2} 
                                                                                          & values=(128,256), milestones=[100]          \\ \hline
    \multirow{2}{*}{\begin{tabular}[c]{@{}c@{}}cutout size\\ (augmentation)\end{tabular}} & 16                                 \\ \cline{2-2} 
                                                                                          & values(16,18,20), milestones=[80,100]     \\ \hline
    optimizer                                                                             & SGD(weight\_decay=4e-5)  \\ \hline
    \end{tabular}
    \caption{Examples from the search space defined for MobileNetV2. 4 types of hyper-parameters were tuned.}
    \label{tbl:mobilenet-searchspace}
\end{table}

\begin{table}[ht]
    \centering
    \begin{tabular}{|c|c|}
    \hline
    \makecell{Hyper\\Parameter}                                                                                  & Function family \\ \hline
    \multirow{2}{*}{learning rate}                                                                   & \makecell{Initial=5e-5,\\ Linear(total\_t=30000)} \\ \cline{2-2} 
                                                                                                     & \makecell{Warmup(3000,5e-5),\\ Linear(total\_t=30000)} \\ \hline
    \multirow{2}{*}{\begin{tabular}[c]{@{}c@{}}input sequence length\\ (preprocessing)\end{tabular}} & 384        \\ \cline{2-2} 
                                                                                                     & \makecell{values=(384,512), \\ milestones=[21000]}   \\ \hline
    \end{tabular}
    \caption{
    Examples from the search space defined for BERT-Base model. 2 types of hyper-parameters were tuned. 
    The warmup and linear decay durations are denoted by number of steps.}
    \label{tbl:bert-searchspace}
\end{table}

\begin{table*}[ht]
    \centering
    \begin{tabular}{|c|c|c|c|c|c|c|c|c|c|c|}
    \toprule
    \multirow{2}{*}{Model}   & \multicolumn{4}{|c|}{Accuracy [\%]} & \multicolumn{3}{|c|}{GPU-Hour} & \multicolumn{3}{|c|}{End-to-End Time [hour]} \\ \cline{2-11}
                             &   Target & Ray Tune & trial & stage & Ray Tune & trial & stage & Ray Tune & trial & stage \\ \hline
    ResNet56 (SHA)           & \multirow{2}{*}{93.03} & 93.08    & 92.89     & 93.27 & 402.66 & 404.95 & 83.7 & 13.92 & 12.89 & 5.76  \\ \cline{1-1} \cline{3-11} 
    ResNet56 (ASHA)          &                        & 93.58    & 92.89     & 93.72 & 544.36 & 374.82 & 139.03 & 17.6 & 13.58 & 7.4   \\ \hline
    MobileNetV2              & 94.43                  & 95.03    & 95.04     & 95.04 & 917.11 & 944.88 & 291.48 & 28.815 & 30.29 & 10.43   \\ \hline
    BERT-Base                & 76.236                 & 78.42    & 78.57     & 78.18 & 835.03 & 808.21 & 404.21 & 25.18 & 24.1 & 11.93    \\ \bottomrule
    \end{tabular}
    \caption{Summary of all four single-study experiments, including the best accuracy reached, elapsed GPU-hour, and end-to-end time. For ResNet56, the target accuracy is the value reached in the original paper. As MobileNetV2 and BERT-Base do not have such official records, their targets were set from values reported in a popular GitHub repository and the SQuAD leaderboard, respectively.}
    \label{tbl:single-study-acc}
\end{table*}

Note that for ResNet56 and MobileNetV2, the reported metric is the top-1 accuracy, and for BERT-Base, the reported metric is the f1 score.
The target top-1 accuracy of ResNet56 is 93.03, which is the value reported in the original paper~\cite{resnet}.
MobileNetV2 has no official record for accuracy on CIFAR-10, but an accuracy of 94.43 is reported in this GitHub repository~\cite{mobilenet-repo}.
The reported f1 score for BERT-Base can be seen on the SQuAD 2.0 official leaderboard, with 76.236 being the highest record for BERT-Base.
The top-1 accuracies and f1 scores reached in our experiments can be found in Table~\ref{tbl:single-study-acc}. In all four \studies, \system successfully achieved top-1 accuracies and f1 scores higher than the reported target values.

The search spaces for ResNet56, MobileNetV2, and BERT-Base are defined in Tables~\ref{tbl:resnet-searchspace}, \ref{tbl:mobilenet-searchspace}, and \ref{tbl:bert-searchspace}, respectively.
The end-to-end time and the GPU-hour are shown in Figure~\ref{fig:single}.
Ray Tune and \system-trial show comparable end-to-end time and GPU-hours except for the study involving ASHA.
In this case, \system-trial finishes earlier and shows a smaller GPU-hour than Ray Tune because a smaller number of trials were promoted due to the asynchronous nature of the algorithm.

Compared to Ray Tune, \system can reduce end-to-end time and GPU-hours by up to 2.76$\times$ and 4.81$\times$, respectively.
For the two grid search experiments, the GPU-hour savings (3.15$\times$ and 2.07$\times$) quite accurately match the value of the merge rate.
This is because the merge rates were calculated assuming that each trial was trained the maximum possible number of iterations, in other words, assuming that a grid search was performed over the search space.
However, for the SHA and ASHA experiments, this assumption does not hold due to early-stopping, and the GPU-hour savings (4.81$\times$ and 3.92$\times$) are much greater than the merge rate.
SHA and ASHA show better performances than anticipated by the search spaces' merge rates because early-stopping lead to exploring only a subset of the whole search space we defined, which happened to have a higher merge rate than the whole search space.
After analyzing the training logs from SHA, for example, we have discovered that the merge rate of the search space actually explored was 4.23.

\subsection{Multiple Studies}

We next evaluate several studies at once, to see the effect of inter-study merging.
We evaluate the GPU-hour and the end-to-end time difference between Ray Tune and \system with a varying number of studies: 1, 2, 4, and 8.
We will refer to each case as S1, S2, S4, and S8.

All studies use the ResNet20 model, the CIFAR-10 dataset, and 144 trials. 
With the \hpoa policy, the merge rate is different for each of the studies. 
Two types of multi-studies are conducted, each with different search spaces. 
The learning rate and batch size were tuned as a sequence in each study. 

\begin{figure*}[ht]
	\centering
	\includegraphics[width=1.0\textwidth]{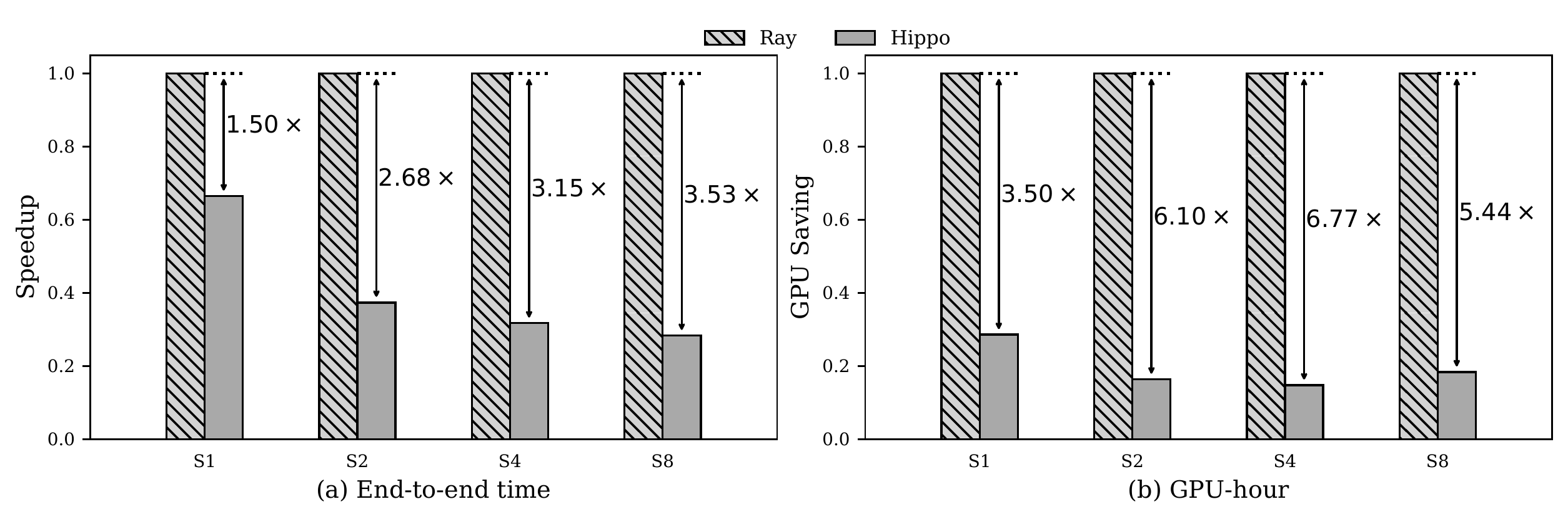}
	\caption{Multi-Study results with k-wise merge rates S2: 2.26, S4: 2.77, and S8: 2.47.}
	\label{fig:multi1}
\end{figure*}

The first search space has comparably higher intra-study and inter-study merge rates. 
The merge rate for each study ranges from 1.5$\times$ to 2.73$\times$.
The k-wise merge rate for S2, S4, and S8 is 2.26, 2.77, and 2.47, respectively.
Figure~\ref{fig:multi1} depicts the results from this search space.
We can see that with a relatively large merge rate between the studies, the GPU-hour and the end-to-end time shrinks by up to 6.77$\times$ and 3.53$\times$. 

\begin{figure*}[ht]
	\centering
	\includegraphics[width=1.0\textwidth]{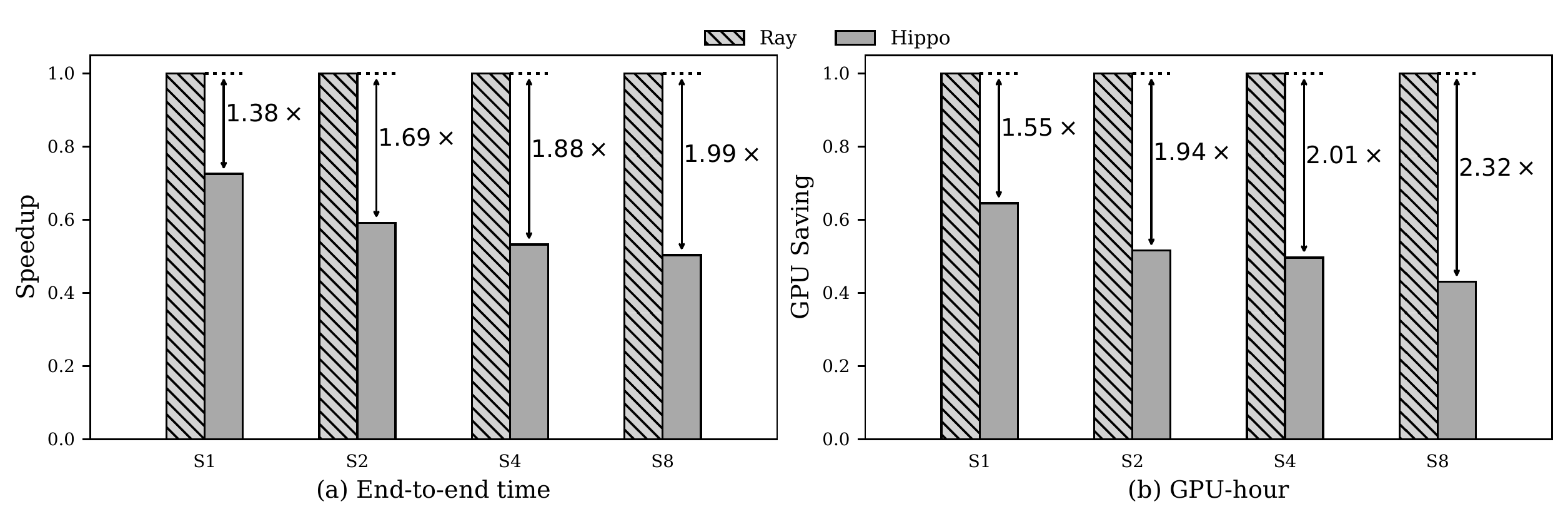}
	\caption{Multi-Study results with k-wise merge rates S2: 1.40, S4: 1.19, and S8: 1.66.}
	\label{fig:multi2}
\end{figure*}

The second search space was designed to have lower intra-study and inter-study merge rates than the first search space.
The merge rate for each study ranges from 1.2$\times$ to 2.1$\times$, and pairwise merge rates also have a similar range of 1.2$\times$ to 2.4$\times$.
The k-wise merge rate for S2, S4, and S8 are 1.40, 1.19, and 1.66, respectively.
Figure~\ref{fig:multi2} depicts the results from this search space.
Though the gains are smaller than in the previously defined search space due to lower merge rates, \system still reduces the GPU-hour and end-to-end time by up to 2.32$\times$ and 1.99$\times$.

\section{Related Work} \label{sec:related}

In our previous workshop paper~\cite{workshop}, we explored the potentials of \stage-based execution by implementing a prototype system and evaluating it with small-scale single study experiments.
Building on these early ideas, in this paper, we present the complete design and implementation details of \system, including challenges and considerations in utilizing \stage-based execution, and added support for multiple studies.
We also provide a richer set of evaluations, consisting of single studies on a much larger scale (in terms of the total number of \trials, types of \hps tuned, and the variety of models and datasets) as well as multi-study experiments, that could better demonstrate the impact of our work.

\paragraph{Trial-based systems}
There have been several recent systems~\cite{raytune, vizier, nni, optuna, Katib, chopt, hyperdrive} for \hpo, helping users to manage their \hpo jobs on distributed environments.
However, the trial-based systems miss the opportunities to reduce resource usage by reusing the common computation results.

Tune~\cite{raytune}, for example, is a \hpo system built on top of Ray~\cite{ray}, providing two-level interfaces: a user API to train models with \hps, and a scheduling API for implementing \hpoas.
Since Tune does not understand the internals of a trial, a single trial cannot be further split into multiple stages to merge the common computation between trials, achieving sub-optimal performance compared to \system.
In addition, since Tune's scheduler is always initiated by the underlying resource manager, e.g., when a trial is completed or a resource is available,
sometimes it may be difficult for a user to manipulate the details of a \hpoa unless the user is familiar with the behavior of the resource manager.
Other popular trial-based \hpo systems such as Google Vizier~\cite{vizier}, NNI~\cite{nni}, Optuna~\cite{optuna}, Kubeflow~\cite{Katib}, CHOPT~\cite{chopt}, and HyperDrive~\cite{hyperdrive} provide similar trial-level user APIs and schedule \hpo jobs in the trial basis, so they also overlook the opportunities to identify and optimize the identical computations as well.

\paragraph{Reusing computation}
\system minimizes resource usage by identifying identical computations and reusing the computation results among multiple hyperparameter optimization trials.
There exist a number of systems that reuse computation results to some extent,
but none of those systems focus on finding out the same computation between \hpo trials as \system does.

Nectar~\cite{nectar} enables reusing common DryadLINQ computation within a datacenter, but does not focus on \hpo jobs.
Pretzel~\cite{pretzel} and Clipper~\cite{clipper} reuse computed results for machine learning inference workloads.
Another recent work~\cite{pipeline-aware-reuse} attempts to reduce the resource usage of \hpo jobs by caching results in intermediate steps of machine learning pipelines such as data preprocessing and feature extraction.
We expect that \system can further be improved by incorporating such techniques that optimize different aspects in machine learning systems.

\paragraph{Systems focusing on a specific algorithm}
As \hpoas such as ASHA~\cite{asha} and PBT~\cite{pbt} have been devised to optimize the resource usage on distributed environments,
systems to efficiently run those algorithms have been introduced alongside with the algorithms themselves.
However, the systems are not generic since each of these systems is specifically designed for executing only a specific algorithm. 
HyperSched~\cite{hypersched} extends ASHA~\cite{asha} and supports algorithms similar to ASHA.
On the other hand, \system targets to support various \hpoas including ASHA~\cite{asha}, SHA~\cite{sha}, PBT~\cite{pbt}, and median-stopping rule~\cite{vizier}, like other existing \hpo systems do.

\section{Conclusion} \label{sec:conc}

\system is a hyper-parameter optimization system that removes redundant computation in the training process by breaking down the hyper-parameter sequences into stages, merging common stages to form a tree of stages, and executing a stage once per tree. \system is applicable to not only single-study scenarios but also multi-study scenarios. Our evaluations show that \system saves GPU-hours and reduces end-to-end training time significantly compared to Ray Tune on multiple models and hyperparameter optimization algorithms. 

\subsubsection*{Acknowledgments}

We thank our colleagues Taebum Kim, Eunji Jeong, Gyeongin Yu, and Won Wook Song for their feedback on this work.
This work was supported by the AWS Machine Learning Research Awards (MLRA), the Institute for Information \& communications Technology Promotion (IITP) grant funded by the Korea government (MSIT) (No.2015-0-00221, Development of a Unified High-Performance Stack for Diverse Big Data Analytics), the ICT R\&D program of MSIT/IITP (No.2017-0-01772, Development of QA systems for Video Story Understanding to pass the Video Turing Test), and Samsung Advanced Institute of Technology.

\bibliographystyle{acm}
\bibliography{hippo}

\begin{thebibliography}{10}

\bibitem{gluster}
{GlusterFS}, 2019.
\newblock \url{https://www.gluster.org/}.

\bibitem{optuna}
{\sc Akiba, T., Sano, S., Yanase, T., Ohta, T., and Koyama, M.}
\newblock Optuna: A next-generation hyperparameter optimization framework.
\newblock In {\em {ACM} {SIGKDD}\/} (2019).

\bibitem{deepspeech2}
{\sc Amodei, D., Ananthanarayanan, S., Anubhai, R., Bai, J., Battenberg, E.,
  Case, C., Casper, J., Catanzaro, B., Cheng, Q., Chen, G., et~al.}
\newblock Deep speech 2: End-to-end speech recognition in english and mandarin.
\newblock In {\em International conference on machine learning\/} (2016),
  pp.~173--182.

\bibitem{hypergradient}
{\sc Baydin, A.~G., Cornish, R., Rubio, D.~M., Schmidt, M., and Wood, F.}
\newblock Online learning rate adaptation with hypergradient descent.
\newblock In {\em International Conference on Learning Representations\/}
  (2018).

\bibitem{survey}
{\sc Bouthillier, X., and Varoquaux, G.}
\newblock Survey of achine-learning experimental methods at {NeurIPS2019} and
  {ICLR2020}.
\newblock {\em Tech report\/} (2020).

\bibitem{gandiva-fair}
{\sc Chaudhary, S., Ramjee, R., Sivathanu, M., Kwatra, N., and Viswanatha, S.}
\newblock {Balancing Efficiency and Fairness in Heterogeneous GPU Clusters for
  Deep Learning}.
\newblock In {\em {EuroSys}\/} (2020).

\bibitem{elu}
{\sc Clevert, D.-A., Unterthiner, T., and Hochreiter, S.}
\newblock Fast and accurate deep network learning by exponential linear units
  (elus), 2015.

\bibitem{clipper}
{\sc Crankshaw, D., Wang, X., Zhou, G., Franklin, M.~J., Gonzalez, J.~E., and
  Stoica, I.}
\newblock {Clipper: A Low-Latency Online Prediction Serving System}.
\newblock In {\em {NSDI}\/} (2017).

\bibitem{mltuner}
{\sc Cui, H., Ganger, G.~R., and Gibbons, P.~B.}
\newblock Mltuner: System support for automatic machine learning tuning, 2018.

\bibitem{imagenet}
{\sc Deng, J., Dong, W., Socher, R., Li, L.-J., Li, K., and Fei-Fei, L.}
\newblock {ImageNet: A Large-Scale Hierarchical Image Database}.
\newblock In {\em CVPR09\/} (2009).

\bibitem{bert}
{\sc Devlin, J., Chang, M.-W., Lee, K., and Toutanova, K.}
\newblock Bert: Pre-training of deep bidirectional transformers for language
  understanding.
\newblock {\em arXiv preprint arXiv:1810.04805\/} (2018).

\bibitem{cutout}
{\sc Devries, T., and Taylor, G.~W.}
\newblock Improved regularization of convolutional neural networks with cutout.
\newblock {\em CoRR abs/1708.04552\/} (2017).

\bibitem{vizier}
{\sc Golovin, D., Solnik, B., Moitra, S., Kochanski, G., Karro, J.~E., and
  Sculley, D.}, Eds.
\newblock {\em Google Vizier: A Service for Black-Box Optimization\/} (2017).

\bibitem{1hour}
{\sc Goyal, P., Doll{\'a}r, P., Girshick, R., Noordhuis, P., Wesolowski, L.,
  Kyrola, A., Tulloch, A., Jia, Y., and He, K.}
\newblock Accurate, large minibatch sgd: training imagenet in 1 hour.
\newblock {\em arXiv preprint arXiv:1706.02677\/} (2017).

\bibitem{tiresias}
{\sc Gu, J., Chowdhury, M., Shin, K.~G., Zhu, Y., Jeon, M., Qian, J., Liu, H.,
  and Guo, C.}
\newblock Tiresias: A {GPU} cluster manager for distributed deep learning.
\newblock In {\em 16th {USENIX} Symposium on Networked Systems Design and
  Implementation ({NSDI} 19)\/} (2019).

\bibitem{nectar}
{\sc Gunda, P.~K., Ravindranath, L., Thekkath, C., Yu, Y., and Zhuang, L.}
\newblock Nectar: Automatic management of data and computation in datacenters.
\newblock In {\em Proceedings of the 9th Symposium on Operating Systems Design
  and Implementation (OSDI)\/} (October 2010).

\bibitem{deepspeech}
{\sc Hannun, A.~Y., Case, C., Casper, J., Catanzaro, B., Diamos, G., Elsen, E.,
  Prenger, R., Satheesh, S., Sengupta, S., Coates, A., and Ng, A.~Y.}
\newblock Deep speech: Scaling up end-to-end speech recognition.
\newblock {\em CoRR abs/1412.5567\/} (2014).

\bibitem{resnet}
{\sc He, K., Zhang, X., Ren, S., and Sun, J.}
\newblock Deep residual learning for image recognition.
\newblock {\em 2016 IEEE Conference on Computer Vision and Pattern Recognition
  (CVPR)\/} (Jun 2016).

\bibitem{rmsprop}
{\sc Hinton, G., Srivastava, N., and Swersky, K.}
\newblock Neural networks for machine learning lecture 6a overview of
  mini-batch gradient descent.

\bibitem{pba}
{\sc Ho, D., Liang, E., Chen, X., Stoica, I., and Abbeel, P.}
\newblock Population based augmentation: Efficient learning of augmentation
  policy schedules.
\newblock In {\em International Conference on Machine Learning\/} (2019),
  pp.~2731--2741.

\bibitem{batchnorm}
{\sc Ioffe, S., and Szegedy, C.}
\newblock Batch normalization: Accelerating deep network training by reducing
  internal covariate shift.
\newblock In {\em Proceedings of the 32nd International Conference on Machine
  Learning\/} (Lille, France, 07--09 Jul 2015), F.~Bach and D.~Blei, Eds.,
  vol.~37 of {\em Proceedings of Machine Learning Research}, PMLR,
  pp.~448--456.

\bibitem{pbt}
{\sc Jaderberg, M., Dalibard, V., Osindero, S., Czarnecki, W.~M., Donahue, J.,
  Razavi, A., Vinyals, O., Green, T., Dunning, I., Simonyan, K., Fernando, C.,
  and Kavukcuoglu, K.}
\newblock Population based training of neural networks, 2017.

\bibitem{sha}
{\sc Jamieson, K., and Talwalkar, A.}
\newblock Non-stochastic best arm identification and hyperparameter
  optimization.
\newblock In {\em Artificial Intelligence and Statistics\/} (2016),
  pp.~240--248.

\bibitem{adam}
{\sc JLB, D. P.~K.}
\newblock Adam: A method for stochastic optimization.
\newblock {\em Proc. of ICLR\/} (2015).

\bibitem{progan}
{\sc Karras, T., Aila, T., Laine, S., and Lehtinen, J.}
\newblock Progressive growing of gans for improved quality, stability, and
  variation, 2017.

\bibitem{chopt}
{\sc Kim, J., Kim, M., Park, H., Kusdavletov, E., Lee, D., Kim, A., Kim, J.-H.,
  Ha, J.-W., and Sung, N.}
\newblock Chopt : Automated hyperparameter optimization framework for
  cloud-based machine learning platforms, 2018.

\bibitem{cifar}
{\sc Krizhevsky, A.}
\newblock Learning multiple layers of features from tiny images.
\newblock Tech. rep., Citeseer, 2009.

\bibitem{Katib}
{\sc kubeflow}.
\newblock Katib: Hyperparameter tuning on kubernetes.

\bibitem{pretzel}
{\sc Lee, Y., Scolari, A., Chun, B.-G., Santambrogio, M.~D., Weimer, M., and
  Interlandi, M.}
\newblock {PRETZEL: Opening the Black Box of Machine Learning Prediction
  Serving Systems}.
\newblock In {\em OSDI\/} (2018).

\bibitem{hyperband}
{\sc Li, L., Jamieson, K., DeSalvo, G., Rostamizadeh, A., and Talwalkar, A.}
\newblock Hyperband: A novel bandit-based approach to hyperparameter
  optimization.
\newblock {\em J. Mach. Learn. Res. 18}, 1 (Jan. 2017), 6765--6816.

\bibitem{asha}
{\sc Li, L., Jamieson, K., Rostamizadeh, A., Gonina, E., Ben-tzur, J., Hardt,
  M., Recht, B., and Talwalkar, A.}
\newblock A system for massively parallel hyperparameter tuning.
\newblock In {\em Proceedings of Machine Learning and Systems 2020\/} (2020),
  pp.~230--246.

\bibitem{massively}
{\sc Li, L., Jamieson, K.~G., Rostamizadeh, A., Gonina, E., Hardt, M., Recht,
  B., and Talwalkar, A.}
\newblock Massively parallel hyperparameter tuning.
\newblock {\em CoRR abs/1810.05934\/} (2018).

\bibitem{pipeline-aware-reuse}
{\sc Li, L., Sparks, E.~R., Jamieson, K.~G., and Talwalkar, A.}
\newblock Exploiting reuse in pipeline-aware hyperparameter tuning.
\newblock In {\em Workshop on Systems for ML at {NeurIPS} 2018\/} (2018).

\bibitem{hypersched}
{\sc Liaw, R., Bhardwaj, R., Dunlap, L., Zou, Y., Gonzalez, J.~E., Stoica, I.,
  and Tumanov, A.}
\newblock Hypersched: Dynamic resource reallocation for model development on a
  deadline.
\newblock In {\em {SOCC}\/} (2019).

\bibitem{raytune}
{\sc Liaw, R., Liang, E., Nishihara, R., Moritz, P., Gonzalez, J.~E., and
  Stoica, I.}
\newblock Tune: A research platform for distributed model selection and
  training.
\newblock {\em arXiv preprint arXiv:1807.05118\/} (2018).

\bibitem{mobilenet-repo}
{\sc liukuang}.
\newblock pytorch-cifar.
\newblock \url{https://github.com/kuangliu/pytorch-cifar}.

\bibitem{themis}
{\sc Mahajan, K., Balasubramanian, A., Singhvi, A., Venkataraman, S., Akella,
  A., Phanishayee, A., and Chawla, S.}
\newblock Themis: Fair and efficient {GPU} cluster scheduling.
\newblock In {\em {NSDI}\/} (2020), pp.~289--304.

\bibitem{nni}
{\sc Microsoft}.
\newblock Neural network intelligence (nni), 2019.

\bibitem{ray}
{\sc Moritz, P., Nishihara, R., Wang, S., Tumanov, A., Liaw, R., Liang, E.,
  Elibol, M., Yang, Z., Paul, W., Jordan, M.~I., and Stoica, I.}
\newblock Ray: A distributed framework for emerging {AI} applications.
\newblock In {\em 13th {USENIX} Symposium on Operating Systems Design and
  Implementation ({OSDI} 18)\/} (Carlsbad, CA, 2018), {USENIX} Association,
  pp.~561--577.

\bibitem{pytorch}
{\sc Paszke, A., Gross, S., Massa, F., Lerer, A., Bradbury, J., Chanan, G.,
  Killeen, T., Lin, Z., Gimelshein, N., Antiga, L., Desmaison, A., Kopf, A.,
  Yang, E., DeVito, Z., Raison, M., Tejani, A., Chilamkurthy, S., Steiner, B.,
  Fang, L., Bai, J., and Chintala, S.}
\newblock Pytorch: An imperative style, high-performance deep learning library.
\newblock In {\em Advances in Neural Information Processing Systems 32},
  H.~Wallach, H.~Larochelle, A.~Beygelzimer, F.~d\textquotesingle
  Alch\'{e}-Buc, E.~Fox, and R.~Garnett, Eds. Curran Associates, Inc., 2019,
  pp.~8024--8035.

\bibitem{hyperdrive}
{\sc Rasley, J., He, Y., Yan, F., Ruwase, O., and Fonseca, R.}
\newblock {HyperDrive: Exploring Hyperparameters with POP Scheduling}.
\newblock In {\em {Middleware}\/} (2017), pp.~1--13.

\bibitem{yolo}
{\sc {Redmon}, J., {Divvala}, S., {Girshick}, R., and {Farhadi}, A.}
\newblock You only look once: Unified, real-time object detection.
\newblock In {\em 2016 IEEE Conference on Computer Vision and Pattern
  Recognition (CVPR)\/} (June 2016), pp.~779--788.

\bibitem{workshop}
{\sc Shin, A., Shin, D.-J., Cho, S., Kim, D.~Y., Jeong, E., Yu, G.-I., and
  Chun, B.-G.}
\newblock Stage-based hyper-parameter optimization for deep learning.
\newblock In {\em Workshop on Systems for ML at {NeurIPS} 2019\/} (2019).

\bibitem{cycliclr}
{\sc Smith, L.~N.}
\newblock Cyclical learning rates for training neural networks.
\newblock {\em 2017 IEEE Winter Conference on Applications of Computer Vision
  (WACV)\/} (Mar 2017).

\bibitem{super}
{\sc Smith, L.~N., and Topin, N.}
\newblock Super-convergence: Very fast training of neural networks using large
  learning rates, 2017.

\bibitem{dont-decay-lr}
{\sc Smith, S.~L., Kindermans, P.-J., and Le, Q.~V.}
\newblock Don't decay the learning rate, increase the batch size.
\newblock In {\em International Conference on Learning Representations\/}
  (2018).

\bibitem{bayesian}
{\sc Snoek, J., Larochelle, H., and Adams, R.~P.}
\newblock Practical bayesian optimization of machine learning algorithms.
\newblock In {\em Advances in Neural Information Processing Systems 25},
  F.~Pereira, C.~J.~C. Burges, L.~Bottou, and K.~Q. Weinberger, Eds. Curran
  Associates, Inc., 2012, pp.~2951--2959.

\bibitem{gnmt}
{\sc Wu, Y., Schuster, M., Chen, Z., Le, Q.~V., Norouzi, M., Macherey, W.,
  Krikun, M., Cao, Y., Gao, Q., Macherey, K., Klingner, J., Shah, A., Johnson,
  M., Liu, X., Kaiser, L., Gouws, S., Kato, Y., Kudo, T., Kazawa, H., Stevens,
  K., Kurian, G., Patil, N., Wang, W., Young, C., Smith, J., Riesa, J.,
  Rudnick, A., Vinyals, O., Corrado, G., Hughes, M., and Dean, J.}
\newblock Google's {Neural} {Machine} {Translation} {System}: {Bridging} the
  {Gap} between {Human} and {Machine} {Translation}.
\newblock {\em arXiv:1609.08144 [cs]\/} (Sept. 2016).
\newblock arXiv: 1609.08144.

\bibitem{gandiva}
{\sc Xiao, W., Bhardwaj, R., Ramjee, R., Sivathanu, M., Kwatra, N., Han, Z.,
  Patel, P., Peng, X., Zhao, H., Zhang, Q., Yang, F., and Zhou, L.}
\newblock Gandiva: Introspective cluster scheduling for deep learning.
\newblock In {\em 13th {USENIX} Symposium on Operating Systems Design and
  Implementation ({OSDI} 18)\/} (Carlsbad, CA, 2018), {USENIX} Association,
  pp.~595--610.

\bibitem{adadelta}
{\sc Zeiler, M.~D.}
\newblock Adadelta: an adaptive learning rate method.
\newblock {\em arXiv preprint arXiv:1212.5701\/} (2012).

\bibitem{yellofin}
{\sc Zhang, J., Mitliagkas, I., and R{\'e}, C.}
\newblock Yellowfin and the art of momentum tuning.
\newblock {\em arXiv preprint arXiv:1706.03471\/} (2017).

\end{thebibliography}

\end{document}